\documentclass[10pt,journal,compsoc]{IEEEtran}
\usepackage[table,dvipsnames]{xcolor}
\usepackage[switch]{lineno}
\usepackage[colorlinks,urlcolor=blue,linkcolor=blue,citecolor=blue]{hyperref}
\usepackage{diagbox}
\usepackage{color,array}
\usepackage{comment}
\usepackage{amsmath}
\usepackage{amssymb}
\usepackage{threeparttable}
\usepackage{graphicx}
\usepackage[export]{adjustbox} 
\usepackage{multirow}
\usepackage{url}
\usepackage{tablefootnote}

\usepackage{lipsum}
\usepackage{soul}
\sethlcolor{yellow}
\usepackage{zref-user,zref-xr}  
\usepackage{xr-hyper}
\usepackage{xr}
\usepackage{cleveref}
\usepackage{algpseudocode}
\usepackage{algorithm}
\usepackage{stackrel}
\usepackage{placeins}

\usepackage{pifont}

\newsavebox{\ORCIDlogo}
\savebox{\ORCIDlogo}{%
\setlength{\unitlength}{\dimexpr 1em/256\relax}%
\begin{picture}(256,256)%
  \color[HTML]{A6CE39}\put(128,128){\circle*{256}}%
  \color{white}%
  \put(78.6,199.2){\circle*{20}}%
  \moveto(70.9,176,9)\lineto(86.3,176,9)\lineto(86.3,69.8)\lineto(70.9,69.8)%
  \closepath\fillpath%
  \moveto(108.9,176.9)\lineto(150.5,176.9)%
  \curveto(190.1,176.9)(207.5,148.6)(207.5 ,123.3)%
  \curveto(207.5,95,8)(186,69.7)(150.7,69.7)%
  \lineto(108.9,69.7)%
  \closepath\fillpath%
  \color[HTML]{A6CE39}%
  \moveto(124.3,83.6)\lineto(148.8,83.6)%
  \curveto(183.7,83.6)(191.7,110.1)(191.7,123.3)%
  \curveto(191.7,144.8)(178,163)(148,163)%
  \lineto(124.3,163)%
  \closepath\fillpath%
\end{picture}%
}
\newcommand\orcidicon[1]{\href{https://orcid.org/#1}{\usebox{\ORCIDlogo}}}
\usepackage{tikz}
\usetikzlibrary{matrix, positioning}
\usepackage{listofitems} 
\usetikzlibrary{arrows.meta} 
\usepackage[outline]{contour} 
\contourlength{1.4pt}
\tikzset{>=latex} 
\colorlet{myred}{red!80!black}
\colorlet{myblue}{blue!80!black}
\colorlet{mygreen}{green!60!black}
\colorlet{myorange}{orange!70!red!60!black}
\colorlet{mydarkred}{red!30!black}
\colorlet{mydarkblue}{blue!40!black}
\colorlet{mydarkgreen}{green!30!black}
\tikzstyle{node}=[thick,circle,draw=myblue,minimum size=22,inner sep=0.5,outer sep=0.6]
\tikzstyle{node in}=[node,green!20!black,draw=mygreen!30!black,fill=mygreen!25]
\tikzstyle{node hidden}=[node,blue!20!black,draw=myblue!30!black,fill=myblue!20]
\tikzstyle{node convol}=[node,orange!20!black,draw=myorange!30!black,fill=myorange!20]
\tikzstyle{node out}=[node,red!20!black,draw=myred!30!black,fill=myred!20]
\tikzstyle{connect}=[thick,mydarkblue] 
\tikzstyle{connect arrow}=[-{Latex[length=4,width=3.5]},thick,mydarkblue,shorten <=0.5,shorten >=1]
\tikzset{ 
  node 1/.style={node in},
  node 2/.style={node hidden},
  node 3/.style={node out},
}
\ifCLASSOPTIONcompsoc
  \usepackage[nocompress]{cite}
\else
  \usepackage{cite}
\fi
\ifCLASSINFOpdf
\else
\fi

\newcommand{\inlinehead}[1]{%
  \par\smallskip\noindent\textbf{#1}\enspace
}

\begin{document}

\title{Graph Attention Based Prioritization of Disease Responsible Genes from Multimodal Alzheimer's Network}


\author{Binon~Teji \orcidicon{0009-0009-4460-044X},~\IEEEmembership{}
    Subhajit~Bandyopadhyay \orcidicon{0000-0003-2272-8935},~\IEEEmembership{}
    Swarup~Roy\textsuperscript{*} \orcidicon{0000-0002-0011-3633}
\IEEEcompsocitemizethanks{
\IEEEcompsocthanksitem \textsuperscript{*}Corresponding Author\protect\\

\IEEEcompsocthanksitem Binon Teji is with Network Reconstruction \& Analysis Lab and Department of Computer Applications, Sikkim University, Sikkim, India, 737102. E-mail: bteji.20pdca01@sikkimuniversity.ac.in \protect\\

\IEEEcompsocthanksitem Subhajit Bandyopadhyay is with Network Reconstruction \& Analysis Lab, Department of Computer Applications, Sikkim University, Sikkim, India, 737102. E-mail: subhajitbn.maths@gmail.com \protect\\

\IEEEcompsocthanksitem Swarup Roy is the corresponding author and with Department of Computer Science \& Engineering, Tezpur University, Assam, India, 784028. E-mail: swarup@tezu.ernet.in \protect\\


}
}

\markboth{Journal of \LaTeX\ Class Files,~Vol.~14, No.~8, August~2015}%
{Shell \MakeLowercase{\textit{et al.}}: Bare Demo of IEEEtran.cls for Computer Society Journals}


\IEEEtitleabstractindextext{%
\begin{abstract}

Prioritizing disease-associated genes is central to understanding the molecular mechanisms of complex disorders such as Alzheimer's disease (AD). Traditional network-based approaches rely on static centrality measures and often fail to capture cross-modal biological heterogeneity. We propose NETRA (Node Evaluation through Transformer-based Representation and Attention), a multimodal graph transformer framework that replaces heuristic centrality metrics with attention-driven relevance scoring. Using AD as a case study, gene regulatory networks are independently constructed from microarray, single-cell RNA-seq, and single-nucleus RNA-seq data. Random-walk sequences derived from these networks are used to train a BERT-based model for learning global gene embeddings, while modality-specific gene expression profiles are compressed using variational autoencoders. These representations are integrated with auxiliary biological networks, including protein-protein interactions, Gene Ontology semantic similarity, and diffusion-based gene similarity, into a unified multimodal graph. A graph transformer assigns NETRA scores that quantify gene relevance in a disease-specific and context-aware manner. Gene set enrichment analysis shows that NETRA achieves a normalized enrichment score of about 3.9 for the Alzheimer's disease pathway, substantially outperforming classical centrality measures and diffusion models. Top-ranked genes enrich multiple neurodegenerative pathways, recover a known late-onset AD susceptibility locus at chr12q13, and reveal conserved cross-disease gene modules. The framework preserves biologically realistic heavy-tailed network topology and is readily extensible to other complex disorders.

\end{abstract}

}

\maketitle
\IEEEraisesectionheading{\section{Introduction}\label{sec:introduction}}

\IEEEPARstart{A}{lzheimer’s} disease (AD) is a progressive neurodegenerative disorder and the most common form of dementia, affecting millions of people around the world\cite{breijyeh2020comprehensive}. Its etiology is very complex and includes a genetic susceptibility intertwined with environmental exposures and lifestyle factors. The precise molecular circuits driving neuronal degeneration remain elusive\cite{sheppard2020alzheimer}. Although high-throughput studies have discovered numerous candidate genes, translating these findings into actionable targets remains a significant challenge. 

From a systems biology perspective, identifying genes associated with AD requires unraveling the intricate molecular interaction networks that drive its progression. This demands accurate reconstruction of gene regulatory networks (GRNs) that closely mirror real biological processes that play an important role in post-inference applications\cite{teji2025gt}. Prioritization of AD genes is a vital task that play crucial roles in disease mechanisms and can aid in small-molecule design for therapeutic interventions. This success is heavily dependent on the precision and completeness of the inferred gene network\cite{sebastian2023generic}. 

Conventional gene prioritization approaches are both resource-intensive and time-consuming, motivating the development of computational alternatives\cite{breijyeh2020comprehensive}. Traditional computational methods largely rely on network-centric measures derived from biological interaction networks, such as degree, betweenness, and eigenvector centrality\cite{sinen2010role}. While these metrics effectively capture topological prominence, they are based on the implicit assumption that highly connected genes are more likely to be disease-relevant. However, in complex diseases such as AD, high network connectivity does not necessarily correspond to functional causality. As a result, centrality-based approaches often fail to capture context-specific gene regulation, modality-dependent expression patterns, and subtle yet biologically meaningful signals that are prevalent in sparse and noisy biological data\cite{sebastian2025network}.


In prior studies on AD gene prioritization, including the work of Wang et al.,\cite{wang2023prioritization} introduces a machine learning framework that leverages spatial–temporal gene expression patterns to classify and predict Alzheimer’s disease risk genes. 
Stephens et al.,\cite{stephens2025computational} propose an integrative GWAS–TWAS framework that prioritizes and functionally validates Alzheimer’s disease risk genes across human and fly cell models. Sebastian et al.~\cite{sebastian2025network} propose an AD gene prioritization framework that analyzes gene positions, AD-specific connectivity patterns, and interaction enrichment within a graph-diffused multimodal regulatory network constructed from transcriptomic, protein–protein interaction (PPI), and Gene Ontology (GO) data, moving beyond reliance on degree centrality alone. Lee et al.~\cite{lee2011towards} highlight that existing gene prioritization methods often rely on single data sources of heterogeneous evidence, and propose a flexible integrative framework that jointly incorporates gene expression, protein–protein interactions, Gene Ontology annotations, and GWAS data to improve Alzheimer’s disease gene prioritization. Cao et al., \cite{cao2025integrating} prioritizes rare, likely deleterious variants from whole-genome sequencing in a Chinese Alzheimer’s disease cohort, identifying novel AD risk genes and functionally linking key variants to cognitive decline, brain atrophy, and amyloid-related cellular dysfunction.

Previous efforts to prioritize genes associated with Alzheimer’s Disease (AD) have largely relied on single-modality data or shallow integration of heterogeneous biological evidence, often exhibiting a bias toward well-studied genes and thereby limiting the discovery of novel candidates. Many approaches fail to capture dynamic gene behavior across tissues, cell types, and experimental conditions. Network-based methods have sought to address these limitations by analyzing topological properties such as node centrality and connectivity; however, such approaches frequently reinforce the assumption that disease-relevant genes are network hubs—an assumption that does not always hold in complex diseases and requires careful validation~\cite{manners2018intrinsic, jha2020prioritizing}. Moreover, existing integrative frameworks typically rely on aggregating independent predictions rather than jointly modeling complementary biological signals, while GWAS–TWAS and experimental validation pipelines face challenges related to scalability, cost, and generalizability. Importantly, the rich and complementary resolution offered by microarray, single-cell RNA sequencing (scRNA-seq), and single-nucleus RNA sequencing (snRNA-seq) data remains underutilized, and many learning-based methods provide limited interpretability of gene relevance.

Motivated by these limitations, we propose a scalable and interpretable gene prioritization framework that jointly integrates heterogeneous transcriptomic, functional, and interaction-based data, rather than combining independent predictions. Our approach moves beyond static network centrality and diffusion-based scores by employing graph attention–based mechanisms that dynamically learn the relative importance of genes in relation to their network neighborhoods. Unlike conventional network analyses that infer disease relevance from fixed topological properties (e.g., degree centrality), graph attention enables a context-aware characterization of gene influence, capturing disease-specific regulatory relationships that may remain obscured under static analysis. By leveraging cross-modal transcriptomic resolution from microarray, scRNA-seq, and snRNA-seq datasets, the proposed framework captures complementary and context-specific disease signals while producing interpretable relevance scores aligned with known AD biology and facilitating the discovery of novel candidate genes—without reliance on extensive experimental validation.


Building on this, we design a unified framework that integrates multi-modal transcriptomic data with regulatory network insights using an attention-guided architecture. Specifically, we infer gene regulatory networks independently from each transcriptomic modality using a suite of algorithms, then embed these networks into rich, contextual graph representations via a BERT-inspired transformer. In parallel, we compress gene expression profiles from each modality into latent vectors using variational autoencoders (VAEs), which are subsequently fused into a consolidated feature space.

A Graph Transformer (GT) module then integrates these network and expression embeddings. The GT leverages positional encodings based on a confidence-weighted ensemble of graph diffusion metrics, Gene Ontology (GO) similarity, and protein-protein interaction (PPI) augmentation. The result is a learned attention score named as \emph{NETRA} for each gene that reflects its biological relevance in the integrated space. This facilitates the prioritization of novel, high-confidence AD candidate genes while offering interpretable attention maps that reveal key regulatory interactions.
By ranking genes according to these attention scores, our method highlights high-confidence AD-associated genes and provides interpretable attention maps that illuminate critical regulatory pathways involved in disease progression.

We validate the biological relevance of \emph{NETRA}-based gene rankings through comprehensive enrichment analyses. Gene Set Enrichment Analysis on KEGG pathways demonstrates that \emph{NETRA} achieves a normalized enrichment score (NES $\approx$ 3.9) for the Alzheimer's disease pathway, substantially outperforming conventional network centrality measures and the SIR diffusion model, which fails to recover Alzheimer's disease among its enriched pathways entirely. Beyond Alzheimer's disease, the top-ranked genes show significant enrichment across multiple neurodegenerative pathways, and analysis of core enrichment genes reveals conserved molecular machinery---particularly cytoskeletal and axonal transport components---shared across Alzheimer's, Parkinson's, Huntington's, Amyotrophic lateral sclerosis, and Prion disease, with pairwise Jaccard similarities reaching up to 0.54. Furthermore, chromosome region enrichment of the top 40 attention-ranked genes identifies a cluster of four genes mapping to the 12q13 cytoband, a locus previously implicated in late-onset Alzheimer's disease through genome-wide association studies, providing independent genomic-level support for the learned gene rankings.

Our contributions are listed as follows: 

\begin{itemize}
    \item We propose a unified and scalable framework that jointly integrates microarray, scRNA-seq, and snRNA-seq data with regulatory and interaction networks for Alzheimer’s disease gene prioritization.

    \item Our method replaces static network centrality and diffusion scores with graph attention mechanisms referred to as \emph{NETRA} that dynamically learn disease-specific gene importance within local and global network contexts.

    \item We introduce a BERT-inspired transformer to embed inferred gene regulatory networks, alongside variational autoencoder–based compression of multi-modal gene expression profiles into a shared latent space.

    \item The framework produces interpretable attention-based gene rankings and regulatory interaction maps, enabling the identification of high-confidence and potentially novel Alzheimer’s disease candidate genes.

    \item We provide multi-level biological validation showing that attention-prioritized genes recover a known AD-associated GWAS locus (chr12q13) and exhibit conserved cross-pathway enrichment across five major neurodegenerative diseases, supporting the biological coherence of the learned rankings.

\end{itemize}


\section{Materials and Methodology}\label{sec:methodology}

\begin{figure*}
    \centering
    \includegraphics[width=0.97\textwidth, height=17.5cm]{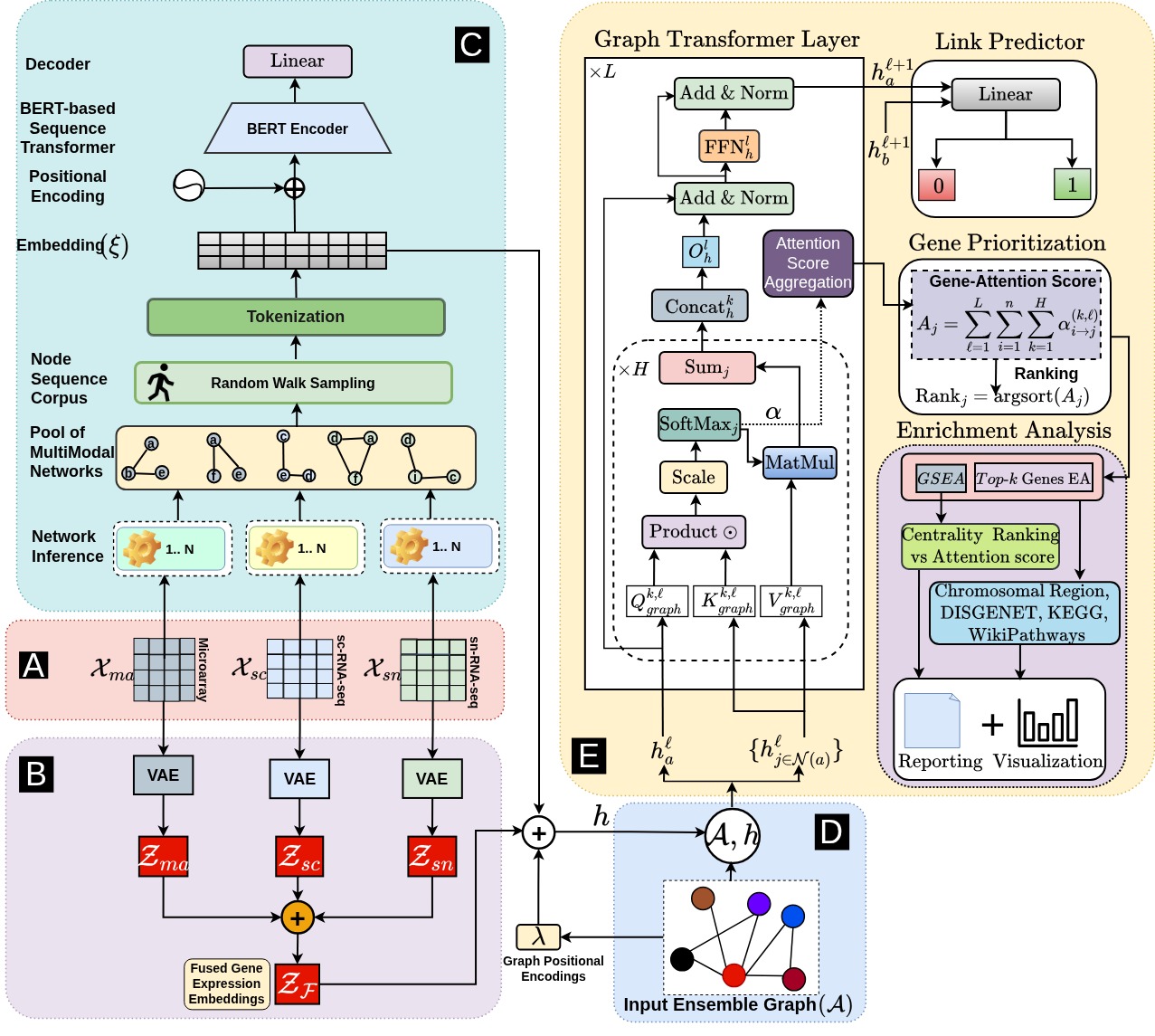}
    \caption{\textbf{Overview of the proposed multimodal graph--transformer framework for gene prioritization.} 
\textit{(A) Multi-omics gene expression data from microarray, single-cell RNA-seq (scRNA-seq), and single-nucleus RNA-seq (snRNA-seq). 
(B) Each modality is encoded using a variational autoencoder (VAE) to obtain latent representations $\mathcal{Z}_{ma}$, $\mathcal{Z}_{sc}$, and $\mathcal{Z}_{sn}$, which are fused into a unified gene embedding $\mathcal{Z}_F$. 
(C) Multiple inferred networks are pooled and transformed into node sequences using random-walk sampling, which are then tokenized, positionally encoded, and encoded using a BERT-based Transformer to learn rich contextual representations.
(D) The fused embeddings are integrated with the ensemble graph $\mathcal{A}$ and graph positional encodings to produce node features $\mathbf{h}$. 
(E) A multi-layer graph transformer applies neighborhood attention using graph-specific query, key, and value projections for link prediction and attention aggregation. The aggregated gene-attention scores are used for gene ranking, enrichment analysis (e.g., GSEA, pathway and disease databases), and final reporting and visualization.}}
    \label{fig:Multi_AD_Prioritization}
\end{figure*}

This section presents the methodology used to prioritize AD genes from multi-modal data, followed by a detailed description of the dataset used. Our framework comprises five core modules: (1) encoding gene expression profiles from multiple modalities; (2) extracting global gene representations through integrated multimodal, multi-network learning; (3) deriving gene positional encodings based on input network structure; (4) fusing these representations to perform gene regulatory network (GRN) inference using a graph transformer; and (5) leveraging attention scores from the model to prioritize AD-relevant genes.
Figure \ref{fig:Multi_AD_Prioritization} illustrates the entire framework. Next, we discuss each module in detail.

\subsection{Variational Autoencoder--Based Multi-Platform Gene Expression Representation Learning}

To capture heterogeneous gene expression patterns across diverse transcriptomic platforms, we employ \textit{modality-specific Variational Autoencoders (VAEs)} for microarray (ma), single-cell RNA sequencing (scRNA-seq), and single-nucleus RNA sequencing (snRNA-seq) data. Each platform exhibits distinct noise characteristics, sparsity patterns, and dynamic ranges, motivating separate encoders while enabling downstream integration through latent feature fusion.

\text{Let} \[
\mathcal{X}_{ma} \in \mathbb{R}^{n_{ma} \times g}, \quad
\mathcal{X}_{sc} \in \mathbb{R}^{n_{sc} \times g}, \quad
\mathcal{X}_{sn} \in \mathbb{R}^{n_{sn} \times g}
\]
denote the gene expression matrices for microarray, scRNA-seq, and snRNA-seq, respectively, where $n$ is the number of genes and $n_e$ represents the number of samples (cells or nuclei) for modality $e \in \{ma, sc, sn\}$. Each modality is modeled using an independent VAE composed of an encoder $E_e$ and a decoder $D_e$. The encoder learns a probabilistic mapping from high-dimensional gene expression space to a compact latent representation:
\[
(\boldsymbol{\mu}_e, \boldsymbol{\sigma}_e) = E_e(\mathcal{X}_e),
\]
which defines the approximate posterior distribution
\[
q(\mathbf{z}_e|\mathcal{X}_e) = \mathcal{N}(\boldsymbol{\mu}_e, \operatorname{diag}(\boldsymbol{\sigma}_e^2)).
\]

This formulation enables the model to capture gene-specific variability, cell-to-cell heterogeneity, and platform-dependent technical noise while enforcing smooth latent representations. Latent variables are sampled using the reparameterization trick:
\[
\mathbf{z}_e = \boldsymbol{\mu}_e + \boldsymbol{\sigma}_e \odot \boldsymbol{\epsilon}, 
\quad \boldsymbol{\epsilon} \sim \mathcal{N}(0, I),
\]
resulting in modality-specific latent embeddings
\[
\mathcal{Z}_{ma}, \quad \mathcal{Z}_{sc}, \quad \mathcal{Z}_{sn}.
\]

These embeddings capture diverse single-gene expression signals, including continuous expression variation (microarray) and sparse, zero-inflated patterns characteristic of single-cell and single-nucleus transcriptomic data. The decoder reconstructs the input expression profiles from the latent space:
\[
\hat{\mathcal{X}}_e = D_e(\mathbf{z}_e),
\]
encouraging the latent representations to preserve biologically meaningful gene expression structure. Each VAE is trained by minimizing the standard evidence lower bound (ELBO):
\[
\mathcal{L}_{\mathrm{VAE}}^{(e)} =
\mathbb{E}_{q(\mathbf{z}_e|\mathcal{X}_e)}
\left[\|\mathcal{X}_e - \hat{\mathcal{X}}_e\|^2\right]
+ \mathrm{KL}\big(q(\mathbf{z}_e|\mathcal{X}_e)\,\|\,\mathcal{N}(0,I)\big),
\]
which balances accurate reconstruction with regularization of the latent space. The learned latent embeddings from all modalities are concatenated to form a unified gene expression representation:
\[
\mathcal{Z}_F = \mathcal{Z}_{ma} \oplus \mathcal{Z}_{sc} \oplus \mathcal{Z}_{sn},
\]

By learning modality-aware yet compact latent representations in a fully unsupervised manner, the proposed VAE framework effectively captures diverse single-gene expression patterns across microarray, scRNA-seq, and snRNA-seq data, enabling robust multi-platform integration for downstream network-based analysis.

\subsection{Global Gene Embedding via Multi-Network Integration of Multi-Modal GRNs}


Although gene expression embeddings capture the intrinsic activity of individual genes, they often lack the contextual or regulatory dependencies that shape biological function. Inferred GRNs provide a complementary view by explicitly modeling both direct and indirect gene–gene interactions. We extend this idea by producing a global embedding for each gene that integrates information from distinct modalities spanning over multiple inferred networks from each modality. 




\subsubsection{Multi-modal Inferred Network Integration via Random Walks and Transformers}

We infer GRNs independently from three expression modalities—microarray, scRNA-seq and snRNA-seq—using a diverse inference algorithms. Each inferred network $G_c = (\mathcal{V}, \mathcal{E}_c)$ (for $c = 1,\dots,C$) shares the same node set $\mathcal{V}$ (genes) but has a modality-specific edge set $\mathcal{E}_c$. The corresponding adjacency matrix $\mathcal{A}^{(c)}\in\{0,1\}^{|\mathcal{V}|\times|\mathcal{V}|}$ encodes these connections, with

$$
\mathcal{A}^{(c)}_{ij} =
\begin{cases}
1, & \text{if } (v_i,v_j)\in\mathcal{E}_c,\\
0, & \text{otherwise}.
\end{cases}
$$

To capture global gene information, we convert each $G_c$ into text-like sequences via random walks (as in node2vec \cite{grover2016node2vec}). Each walk of length $ln$ yields a sequence of node tokens $ [v_{w_1}, v_{w_2}, \dots, v_{w_{ln}}]$, which we prepend with a special $[CLS]$ token. We embed these tokens through a shared embedding matrix $\xi\in\mathbb{R}^{|\mathcal{V}|\times d_n}$, where $d_n$ is the network-embedding dimension. To retain positional information, we add sinusoidal position encodings $PE_{\text{seq}}(\text{pos},i)$ as done in \cite{vaswani2017attention}. The sum of token embeddings and positional encodings for all walks across all $C$ networks forms the input to a transformer model, whose parameters are learned jointly—in particular, using Xavier initialization for the transformer weights \cite{devlin2019bert}. During training, nodes are randomly masked in each sequence, and the model learns to predict these masked tokens, encouraging it to capture higher-order network context. The final hidden state of the $[CLS]$ token for each sequence serves as the global embedding for the corresponding gene, now informed by multi-modal network structure as well as its original expression context.


\subsubsection{Masked Language Modeling via BERT for Multi-Modal GRN Corpus}

We treat each inferred gene regulatory networ constructed independently from microarray, scRNA-seq, and snRNA-seq expression data—as a source of contextual gene relationships and jointly organize them into a unified multi-modal corpus. To learn contextualized gene embeddings from this corpus, we adopt the Masked Language Modeling (MLM) objective from BERT~\cite{devlin2019bert} and employ a transformer encoder with $M$ stacked self-attention layers~\cite{vaswani2017attention}. Random walks sampled from each modality-specific GRN are represented as sequences of gene tokens,
\[
F = [f_1, f_2, \dots, f_n],
\]
where each token corresponds to a gene and is mapped to a $p$-dimensional embedding. These sequences are processed by the transformer encoder, which models long-range and higher-order regulatory dependencies among genes through self-attention. Since the transformer architecture and attention mechanism are standard, we omit their full formulation for brevity.

To enable self-supervised learning, we randomly mask $20\%$ of gene tokens in each sequence—across all modalities—and train the model to predict the masked genes using the remaining contextual information. The training objective is the standard cross-entropy loss applied only to masked positions:

\begin{equation}
L_{MLM} = -\sum_{b=1}^{B} \sum_{ln=1}^{Ln} \sum_{cl=1}^{Mc} 1_{\{b, ln \in \text{mask}\}} \cdot y_{b,ln,cl} \log p_{b,ln,cl},
\end{equation}

where $\mathcal{M}$ denotes the set of masked token positions, $|\mathcal{V}|$ is the gene vocabulary size, $y_{b,ln,cl}$ is the ground-truth indicator, and $p_{b,ln,cl}$ is the predicted probability for gene $cl$.

After training, the learned gene representations are obtained from the shared embedding matrix,
\[
\boldsymbol{\xi} \in \mathbb{R}^{|\mathcal{V}| \times p},
\]
where each row of $\boldsymbol{\xi}$ corresponds to a global gene embedding. These embeddings integrate regulatory information from multiple inferred GRNs and transcriptomic modalities, capturing both shared and modality-specific gene interaction patterns within a single unified space.

The resulting embeddings $\boldsymbol{\xi}$ serve as the global gene representations used in subsequent attention-based integration and disease-specific gene prioritization tasks.


\subsection{GRN Inference via Graph Transformer}

Subsequent to extracting quantitative embeddings from multi-modal expression data and capturing structural insights from multiple inferred GRNs, it becomes crucial to integrate these complementary sources into a coherent model to understand gene regulation. To achieve this, we employ a Graph Transformer (GT)\cite{dwivedi2020generalization} architecture, which is specifically designed to handle graph-structured input via learnable attention mechanisms. By supplying the GT with both multi-modal expression embeddings and inferred network structures, the model can jointly attend to key regulatory signals and interaction patterns, enabling robust GRN reconstruction. This integration produces concise, context-aware gene embeddings that capture both the underlying network topology and functional heterogeneity across modalities. Next, we discuss about the inner workings of the GT method.

\subsubsection{Graph Positional Encodings}
Similar to the role of positional encodings in sequence-based Transformers\cite{vaswani2017attention}. At the core of GT architecture, GT relies on graph positional encodings ($PE_{graph}$) to incorporate node-wise structural context. We take advantage of the spectral characteristics of the input network. We compute the eigenvectors of the graph Laplacian, which act as position-aware features for nodes. These encodings allow the model to account for graph topology, enabling it to distinguish between local and global relationships between nodes in a structure-informed manner.

Given an input adjacency matrix $\mathcal{A} \in \mathbb{R}^{n \times n}$, we compute the graph Laplacian matrix:

\begin{equation}
\Delta = I_{graph} - D^{-\frac{1}{2}} \mathcal{A} \thinspace D^{-\frac{1}{2}} = U \Lambda U^{T}
\end{equation}

where $I_{graph}$ is the identity matrix, $D$ is the degree matrix, $\Lambda$ contains the eigenvalues, and $U$ comprises the corresponding eigenvectors. The $p$ smallest non-trivial eigen-vectors from $U$ are selected as $PE_{graph}$ for each node, denoted by $\lambda_i$. 

These encodings are particularly useful for modeling distance-aware relationships. Nodes that are close in the network tend to have similar positional embeddings, while distant nodes are represented more distinctly.
By integrating $PE_{graph}$ into the GT pipeline, the model can effectively encode both topological context and functional diversity, enabling richer and more structure-informed representation for a better GRN reconstruction.

\subsubsection{Confident Gene Network Construction}
To build a robust, integrative representation of Alzheimer’s disease (AD) gene regulation, we employ the high-confidence consensus network developed by Sebastian et al. \cite{sebastian2025network}. This consensus graph forms the structural backbone of the Graph Transformer (GT) model, which is further enriched with pre-trained multi-modal expression embeddings and global gene features derived from BERT for robust GRN inference and further aid in gene prioritization. Primarily, authors curate and preprocess three complementary transcriptomic datasets from bulk microarray, single-cell RNA-seq, and single-nucleus RNA-seq to generate noise-reduced, normalized expression matrices. These are enriched with human protein–protein interaction scores from STRING and Gene Ontology similarity priors, each scaled to [0, 1]. For each modality, eight diverse network-inference algorithms (CLR, MutRank, MINE, MRNETB, WGCNA, PCOR, SPCOR, LEAP) produce six microarray, four scRNA-seq, and three snRNA-seq GRNs, alongside PPI and GO–based graphs. Rather than simply averaging these fifteen undirected, weighted networks, they apply a graph-diffusion ensemble to repeatedly propagate and reinforce edge weights to amplify consistently supported interactions and suppress noise. This yields a single, robust consensus network of 11229x11229 dimension. This unified network then supports GT-based GRN modeling and attention-driven prioritization of AD-related genes.

\subsubsection{Input to the Graph Transformer}
The Graph Transformer (GT) receives as input the consensus graph structure \(\mathcal{G} = (\mathcal{V}, \mathcal{E})\) and a feature matrix \(h^{(0)} \in \mathbb{R}^{n \times d}\), where \(n = |\mathcal{V}|\) and \(d\) is the length of the each gene vector. Each row \(h_i^{(0)}\) is the sum of three projected embeddings for gene \(i\):

\begin{equation}
    h_i^{(0)} = \underbrace{S^0 \mathcal{Z}_i + s^0}_{\text{expression}} \;+\; \underbrace{T^0 \,\xi_i + t^0}_{\text{global}} \;+\; \underbrace{U^0 \,\lambda_i + u^0}_{\text{positional}},
\end{equation}

where, \(\mathcal{Z}_i \in \mathbb{R}^{d_n}\) is the multi-modal expression embedding, projected by \(S^0\in\mathbb{R}^{d\times d_n}\) and bias \(s^0\in\mathbb{R}^d\). \(\xi_i \in \mathbb{R}^{d_n}\) is the global BERT-derived embedding, projected by \(T^0\in\mathbb{R}^{d\times d_n}\) and bias \(t^0\in\mathbb{R}^d\). \(\lambda_i \in \mathbb{R}^{d_n}\) is the graph positional encoding from Laplacian eigenvectors, projected by \(U^0\in\mathbb{R}^{d\times d_n}\) and bias \(u^0\in\mathbb{R}^d\). 

The initial feature vector for each gene $i$, denoted $h_i^{(0)}$, is constructed by fusing three complementary embeddings, i.e., multi-modal gene expression embddings, global gene embeddings, and gene positional encodings derived from input concensus network. By summing these three projected vectors, $h_i^{(0)}$ jointly represents each gene’s expression dynamics, global context, and network topology. This rich, unified embedding is then passed along with the adjacency matrix $\mathcal{A}$ into the Graph Transformer layer.

\subsubsection{Graph Transformer Layer}

At layer $\ell$, each node $i$ attends over its local neighborhood $\mathcal{N}(i)$ using a multi-head graph attention mechanism.
Let $h_i^{(\ell)} \in \mathbb{R}^d$ denote the node representation at layer $\ell$.
The attention-based aggregation is defined as
\begin{equation}
\hat{h}_{i}^{\ell+1}
=
O_h^{\ell}
\bigg\|
_{k=1}^{H}
\sum_{j \in \mathcal{N}(i)}
\alpha_{ij}^{k,\ell}
V_{graph}^{k,\ell} h_j^{\ell},
\end{equation}
where $\|$ denotes concatenation over $H$ attention heads and $O_h^{\ell} \in \mathbb{R}^{d \times d}$ is the output projection matrix.
The attention coefficients $\alpha_{ij}^{k,\ell}$ are computed using scaled dot-product attention,
\begin{equation}
\alpha_{ij}^{k,\ell}
=
\mathrm{softmax}_j
\left(
\frac{
Q_{graph}^{k,\ell} h_i^{\ell}
\cdot
K_{graph}^{k,\ell} h_j^{\ell}
}{\sqrt{d_k}}
\right),
\end{equation}
where $Q_{graph}^{k,\ell}$, $K_{graph}^{k,\ell}$, and $V_{graph}^{k,\ell} \in \mathbb{R}^{d_k \times d}$ are learnable projection matrices, $d_k = d/H$, and $H$ denotes the number of attention heads.

The aggregated representation $\hat{h}_i^{\ell+1}$ is then passed through a position-wise feed-forward network (FFN) with residual connections and normalization, yielding the final node representation:


\begin{equation}
\begin{aligned}
h_i^{\ell+1}
&=
\mathrm{Norm}\Big(
\hat{h}_i^{\ell+1}
+
W_{graph2}^{\ell}
\,\mathrm{ReLU}\Big(
W_{graph1}^{\ell}
\\
&\qquad\qquad
\cdot
\mathrm{Norm}\big(
h_i^{\ell}
+
\hat{h}_i^{\ell+1}
\big)
\Big)
\Big)
\end{aligned}
\end{equation}

Here, $W_{graph1}^{\ell} \in \mathbb{R}^{2d \times d}$ and $W_{graph2}^{\ell} \in \mathbb{R}^{d \times 2d}$ are the FFN weight matrices.
The normalization operator $\mathrm{Norm}(\cdot)$ corresponds to either Layer Normalization~\cite{ba2016layer} or Batch Normalization~\cite{ioffe2015batch}.
This formulation enables effective aggregation of neighborhood information while preserving stability and expressive power through residual connections and normalization.

\subsubsection{Network Reconstruction via Link Prediction }
This component predicts the presence of edges using the node/gene embeddings $h^{(\ell+1)} \in \mathbb{R}^{n \times d}$ produced by the GT. Given the matrix $h^{(\ell+1)} \in \mathbb{R}^{n\times d}$ and a list of candidate node pairs, we extract each pair’s embeddings $h^{(\ell+1)}_i$ and $h^{(\ell+1)}_j$, concatenate them into a single feature vector, and feed this into a decoder multi-layer perceptron (MLP). The decoder consists of one hidden layer with ReLU activation followed by an output layer that maps the combined features to a single scalar score. This score yields the predicted probability of an edge between nodes $i$ and $j$. By leveraging the structure and context-rich embeddings from the GT layers, this module effectively reconstructs the input graph into a matrix of predicted edge probabilities, enabling accurate recovery of the underlying regulatory interactions.

\subsubsection{\emph{NETRA} for Attention-Based Gene Prioritization}
Once the Graph Transformer (GT) has been trained using the consensus gene-gene interaction network, which is enriched with fused gene representations comprising multi-modal expression embeddings, global BERT-derived gene embeddings, and positional gene encodings extracted from the concensus network. It reconstructs the consensus network, facilitating potential GRN inference. Beyond GRN inference, the learned attention mechanisms and influence scores derived from GT can be ultimately leveraged towards prioritizing AD-related genes with potential biological significance. GT achieves this by applying multi-head self-attention over the network structure, effectively capturing complex dependencies among genes.

For each GT layer $\ell = 1, 2, \dots, L$, let $\alpha_{i \to j}^{(k, \ell)}$ denote the attention weight from the source node $i$ to the target node $j$ for head $k = 1, \dots, H$. We first aggregate across heads:

\begin{equation}
    \alpha_{i \to j}^{(\ell)} = \sum_{k=1}^H \alpha_{i \to j}^{(k, \ell)}
\end{equation}

Next, we compute the total incoming attention for node $j$ in layer $\ell$:

\begin{equation}
    A_j^{(\ell)} = \sum_{i=1}^n \alpha_{i \to j}^{(\ell)}
\end{equation}

Finally, we compute the overall influence score $A_j$ by summing over all layers and heads:

\begin{equation}
    A_j = \sum_{\ell=1}^L \sum_{i=1}^n \sum_{k=1}^H \alpha_{i \to j}^{(k, \ell)}
\end{equation}

where, $A_j$ represents the overall influence score for node $j$ (i.e., a gene). This score reflects the importance or centrality of gene $j$ in the concensus graph, as determined by the model's learned attention weights. Similarly, we obtain a ranked list of genes based on their aggregated attention scores. This ranking highlights genes that play influential roles in the regulatory landscape, enabling systematic gene prioritization.

Next, we discuss the experimental setup used to demonstrate the overall framework.

\subsection{Experimental Setup}
The performance of our framework is evaluated on an NVIDIA RTX A3000 GPU.  
The deep-learning libraries used are PyTorch\footnote{\href{https://pytorch.org/}{https://pytorch.org/}},  
DGL\footnote{\href{https://www.dgl.ai/}{https://www.dgl.ai/}},  
scikit-learn\footnote{\href{https://scikit-learn.org/stable/}{https://scikit-learn.org/stable/}},  
and PyTorch Geometric\footnote{\href{https://pytorch-geometric.readthedocs.io/en/latest/index.html}{https://pytorch-geometric.readthedocs.io/}}

\subsection{Multi-Modal Data Collection}

We retrieved the microarray data with GEO Accession \#GSE1297\cite{blalock2004incipient} along with the scRNA-seq dataset with GEO accession \#GDS1979\cite{muller2007modulation} from the GEO database\footnote{\url{https://www.ncbi.nlm.nih.gov/geo/}}. Further, the snRNA-seq dataset with scREAD
Data ID: AD00204 is procured from scREAD: a single-cell RNA-Seq database
for Alzheimer’s disease\cite{jiang2020scread}

\begin{table*}[!ht]
\centering
\caption{Summary of the characteristics of the multiomics datasets.}
\begin{adjustbox}{height = 0.6in, width=5.25in}
\begin{tabular}{|l|l|l|l|l|l|l|}
\hline

\begin{tabular}[c]{@{}c@{}}
\textbf{Dataset} \\ \textbf{Type}
\end{tabular}
 & \textbf{Source} & \textbf{Accession \#} & \textbf{Organism} & 
    \begin{tabular}[c]{@{}c@{}}
    \textbf{No. of} \\ \textbf{Genes}   
    \end{tabular}
     
  &
  \begin{tabular}[c]{@{}c@{}}
  \textbf{Tissue/} \\
  \textbf{Cell Type}
  \end{tabular}
   &
   \begin{tabular}[c]{@{}c@{}}
   
   \textbf{No. of} \\ \textbf{Samples}
   \end{tabular}
    \\
\hline
Microarray & GEO & GSE1297 & \textit{Homo sapiens} & 23000 & 

   \begin{tabular}[c]{@{}c@{}}

hippocampal\\ gene expression 
\end{tabular}
& 31 \\
\hline
Microarray      & GEO & GDS1979 & \textit{Homo sapiens} & 23000 & neural cells                & 6  \\
\hline
scRNA      & GEO & GSE129308 & \textit{Homo sapiens} & 23000 & neural cells                & 27  \\
\hline
snRNA      & scREAD & AD00204 & \textit{Homo sapiens} & 22000 & neurons                     & 201 \\

\hline
\end{tabular}
\end{adjustbox}
\end{table*}

\section{Results}
In this section, we first evaluate the quality of the learned gene representations and the structural validity of the generated network. We then analyze interaction patterns and topological properties of attention-prioritized genes, followed by an assessment of their biological relevance through functional enrichment analysis.



\subsection{Embedding and Network-Level Validation of Attention-Based Prioritized Genes}

\begin{figure}[t]
    \centering
    \includegraphics[width=1.02\linewidth, height=4.5cm]{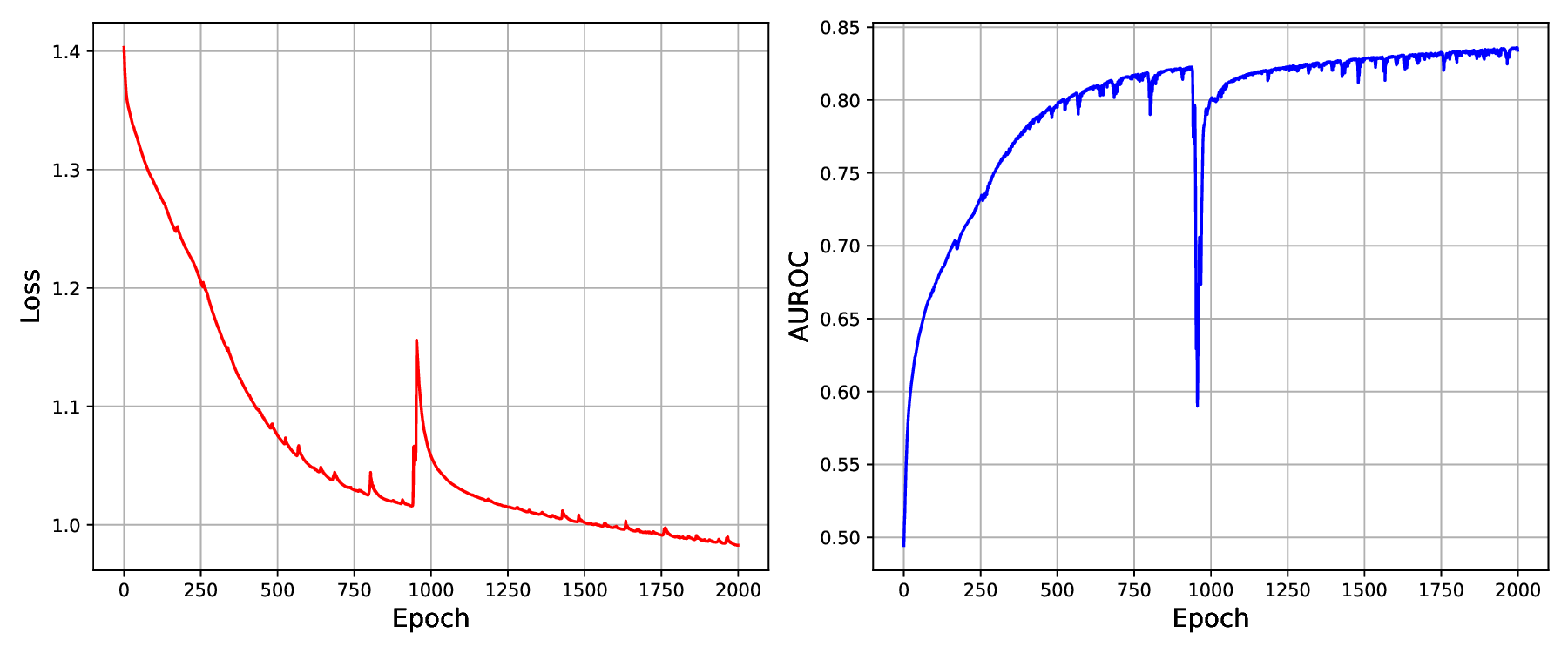}
    \caption{{\bf Training dynamics of our learning model.} \emph{(left) training loss convergence and (right) AUROC progression across epochs, demonstrating stable optimization and robust generalization.}}
    \label{fig:auroc_epochs}
\end{figure}

\begin{figure}[t]
    \centering
    \includegraphics[width=\linewidth]{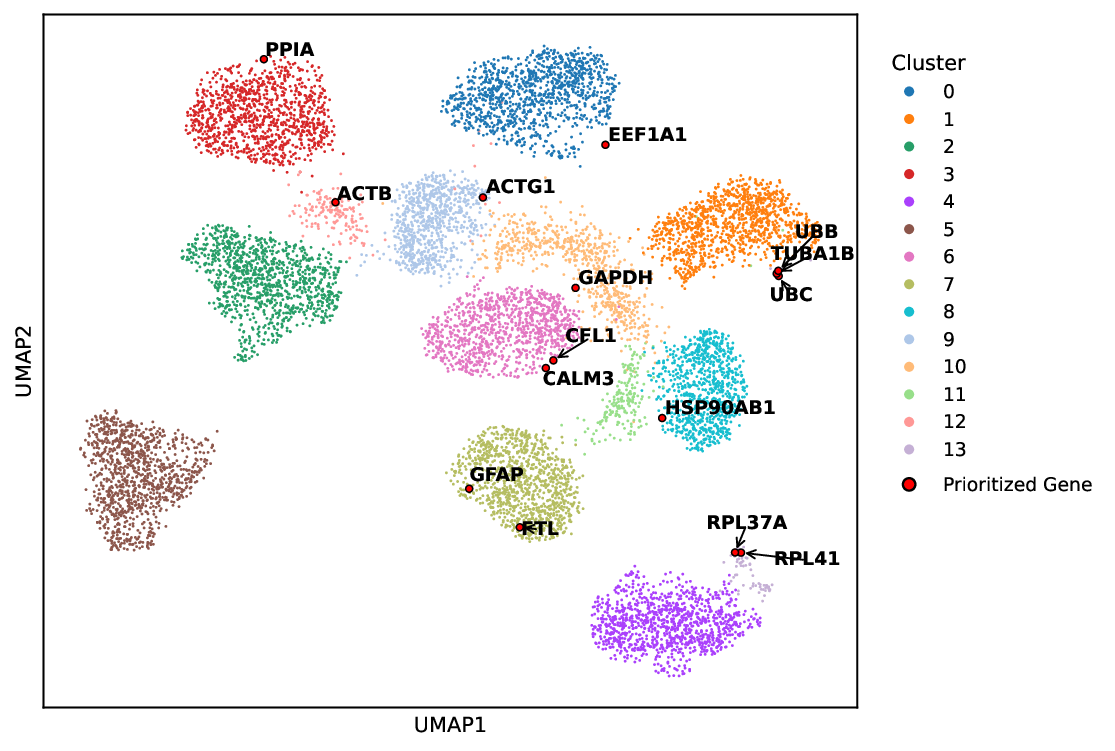}
    \caption{\textbf{UMAP visualization of gene embeddings with Leiden clustering and attention-prioritized (top-15) genes.}
    Each point represents a gene embedding projected into two dimensions. Colors indicate Leiden clusters (labels 0–13).
    Attention-prioritized genes are highlighted in red and annotated with gene symbols using arrowed labels.
    The spatial distribution demonstrates clear cluster separation while showing the dispersion of prioritized genes across multiple functional communities.}
    \label{fig:umap_leiden_prioritized}
\end{figure}

\begin{figure*}[t]
    \centering
    \includegraphics[width=\textwidth]{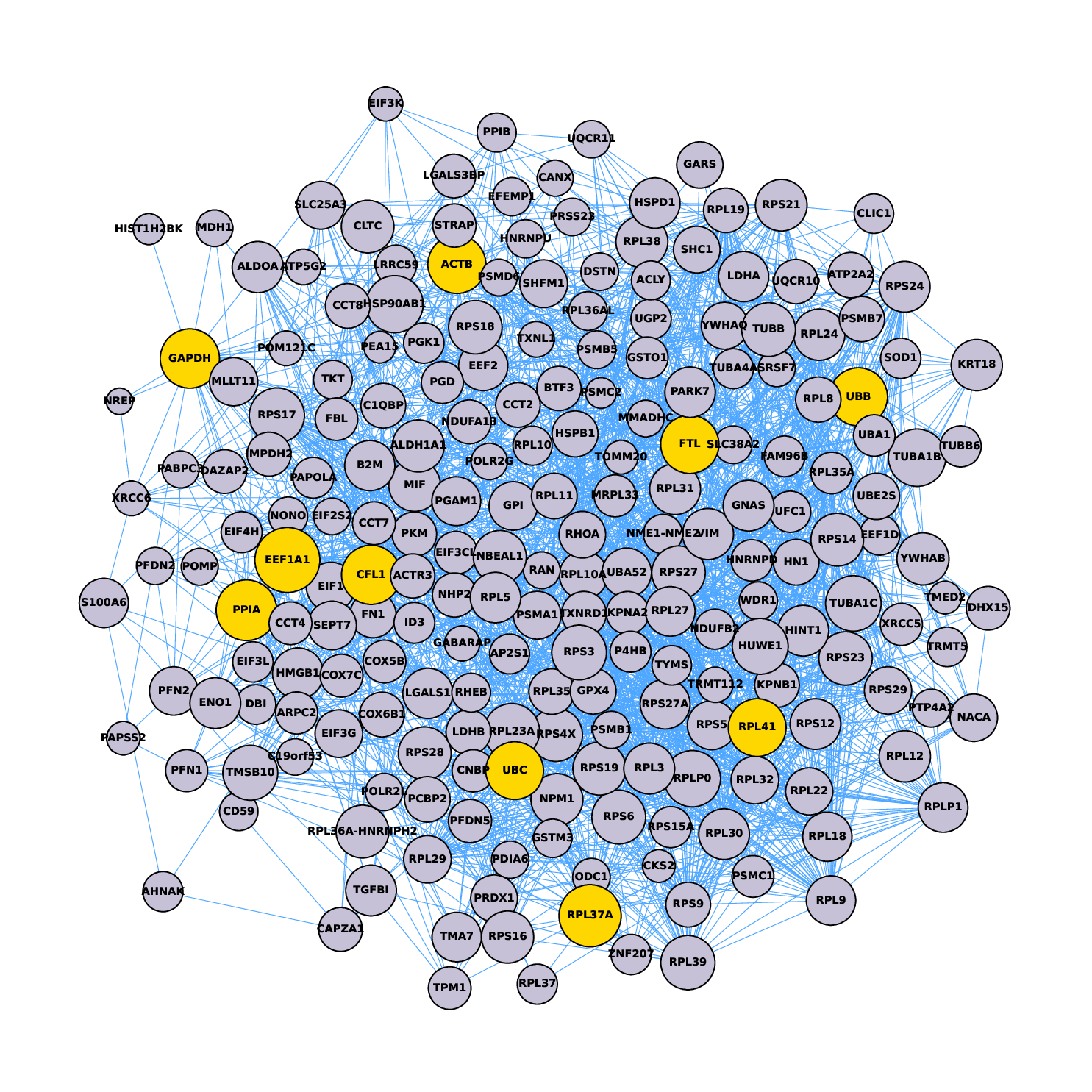}
    \caption{{\bf Visualization of an attention-weighted subgraph induced by the top-10 prioritized genes.} \emph{Prioritized genes are shown in gold, neighboring context genes in gray, with node sizes proportional to \textit{NETRA} attention scores. Blue edges represent gene-gene interactions.}}\label{fig:subgraph}
\end{figure*}

We first examine the training dynamics of the proposed framework during the network reconstruction phase. Fig.~\ref{fig:auroc_epochs} illustrates the evolution of training loss and validation AUROC across epochs. The consistent decrease in loss accompanied by a steady increase in AUROC indicates stable convergence and effective learning of discriminative gene representations. Moreover, the saturation of AUROC in later epochs suggests that the model achieves good generalization performance without overfitting. Then we evaluate the quality of the learned gene representations produced by the proposed attention-based framework. To assess whether the model captures meaningful structural organization, we projected the generated gene embeddings into a two-dimensional space using UMAP and applied Leiden community detection. As shown in Fig. \ref{fig:umap_leiden_prioritized}, the embedding space exhibits well-defined clusters, indicating that the model learns structured and coherent representations rather than a diffuse or random embedding. These clusters correspond to distinct embedding communities, suggesting the presence of modular organization among genes in the generated network. Importantly, attention-prioritized genes are not confined to a single cluster, but are distributed across multiple Leiden communities. This dispersion indicates that the prioritization mechanism does not merely select genes from one dense region of the embedding space or favor high-degree hubs, but instead identifies genes spanning diverse functional modules. Arrow-based annotations highlight representative prioritized genes within and across clusters, further illustrating the ability of the model to capture heterogeneous but biologically meaningful gene groups. To examine the interaction patterns among highly ranked genes, we extracted the induced subgraph of the top-50 attention-prioritized genes from the generated network (Fig. \ref{fig:subgraph}). In this attention-weighted subnetwork, nodes represent genes and edges denote inferred interactions learned by the model. Node sizes are scaled according to $log2$ transformed attention scores, emphasizing genes that contribute most strongly to the model’s prioritization decision. The resulting subgraph reveals dense connectivity and tightly coupled modules among top-ranked genes, suggesting coordinated functional relationships rather than isolated gene selection. Such modular structure is consistent with the expectation that disease-relevant genes participate in interconnected biological pathways, particularly in complex disorders such as Alzheimer’s disease.Beyond local connectivity patterns, we assessed whether the generated network preserves global topological properties of the input ensemble network. 

\begin{figure*}[!t]
    \centering

    \includegraphics[width=0.4\textwidth, height=5.45cm]{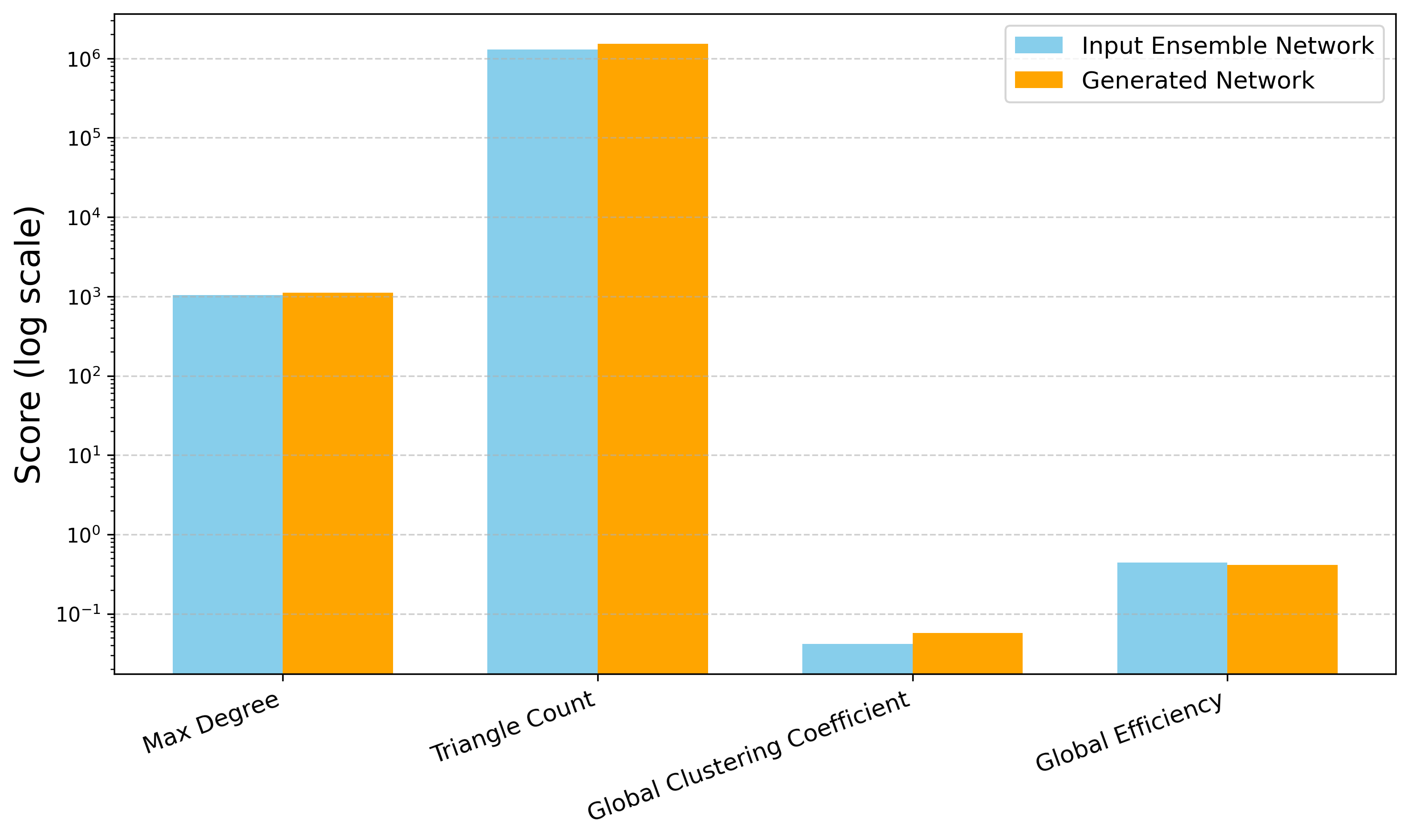} \hspace{0.6cm}
    \includegraphics[width=0.4\textwidth]{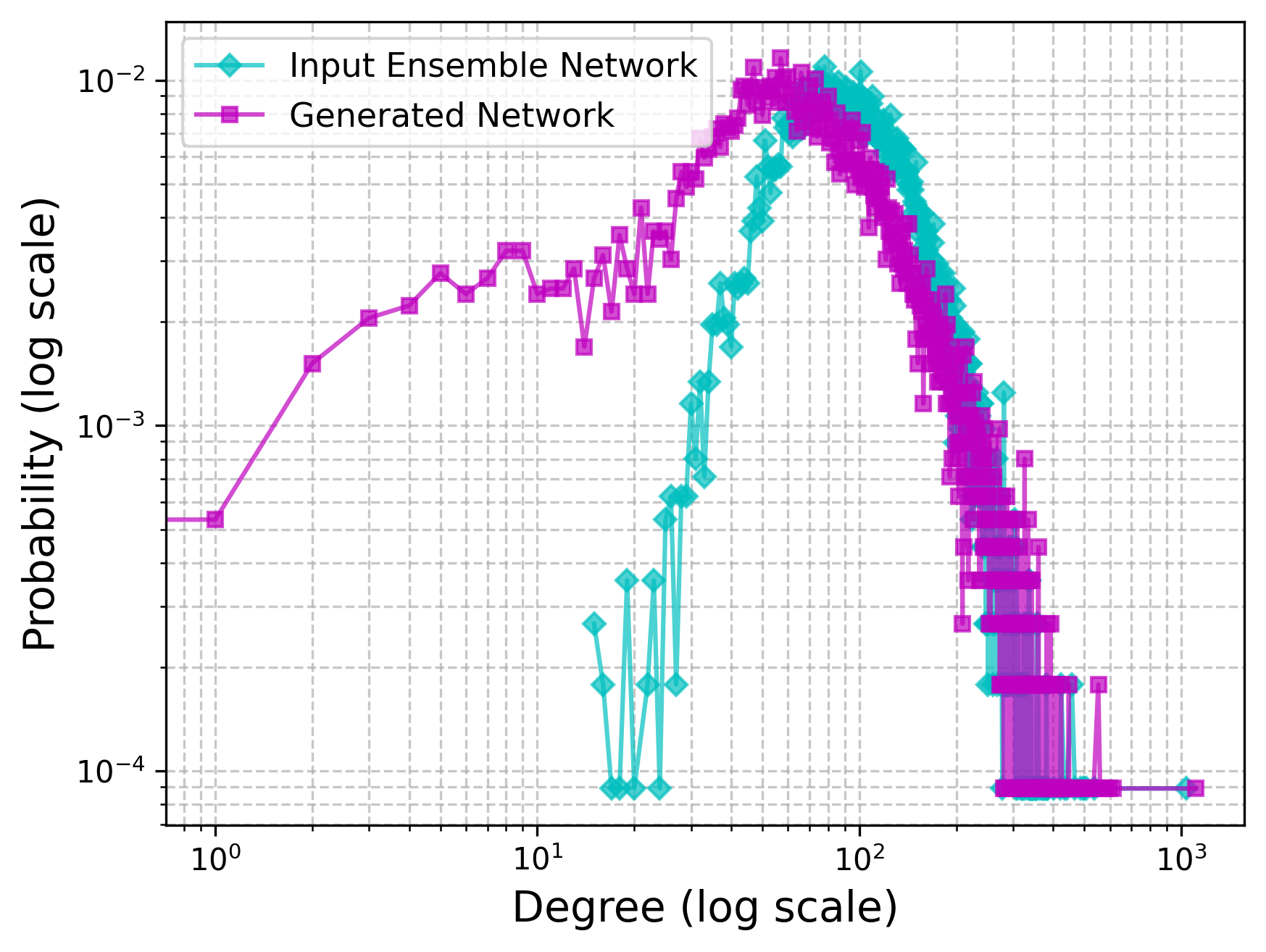}\\
    \caption{{\bf Structural comparison between the input ensemble network and the generated network.} \emph{(a) Comparison of global graph characteristics between the input ensemble network and the generated network on a logarithmic scale, including maximum degree, triangle count, global clustering coefficient, and global efficiency. (b) Log–log degree distribution of the input ensemble network and the generated network, demonstrating the preservation of heavy-tailed degree behavior in the generated network.}}\label{fig:structural_comparison}
\end{figure*}

Figure \ref{fig:structural_comparison} compares key graph characteristics between the two networks, including maximum degree, triangle count, global clustering coefficient, and global efficiency, all evaluated on a logarithmic scale. The generated network closely matches the input ensemble across these metrics, indicating that the generative process maintains both local clustering tendencies and global communication efficiency. This structural consistency suggests that the learned network is not an artifact of overfitting to a small subset of nodes, but rather reflects realistic network organization. To further support for this observation we plot the degree distribution of the input and the generated network. Both the input ensemble network and the generated network exhibit heavy-tailed degree distributions on a log–log scale, a hallmark of biological interaction networks. The preservation of this property demonstrates that the generated network retains scale-free characteristics, reinforcing the validity of the learned topology. Taken together, these results demonstrate that the proposed framework learns structured gene embeddings, prioritizes genes across diverse embedding communities, and generates a network that faithfully preserves essential topological properties of the original ensemble. Having established the embedding-level coherence, network-level realism, and non-trivial distribution of prioritized genes, we next investigate the biological relevance of the attention-ranked gene set through functional enrichment analysis.

\subsection{Biological Validation of Results}
To validate the biological relevance of attention-based gene prioritization, we performed comprehensive enrichment analyses at two levels: (1) pathway-level Gene Set Enrichment Analysis (GSEA) comparing attention scores against network centrality measures, and (2) targeted enrichment of top-ranked genes across pathways, diseases, and chromosomal regions.

\subsubsection{Gene Set Enrichment Analysis: Attention Scores vs. Network Centrality}
We evaluated whether attention-derived rankings (\emph{NETRA}) better capture Alzheimer's disease biology compared to traditional network centrality measures (degree, betweenness, eigenvector, and PageRank) using Gene Set Enrichment Analysis on KEGG pathways.

\begin{figure*}[!t]
    \centering

    {\includegraphics[width=0.40\textwidth]{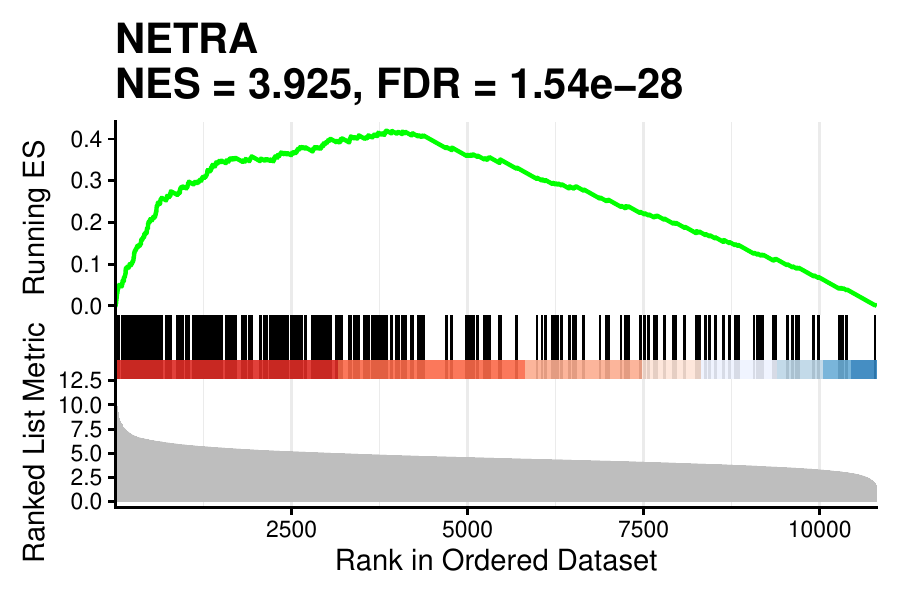}}
    {\includegraphics[width=0.40\textwidth]{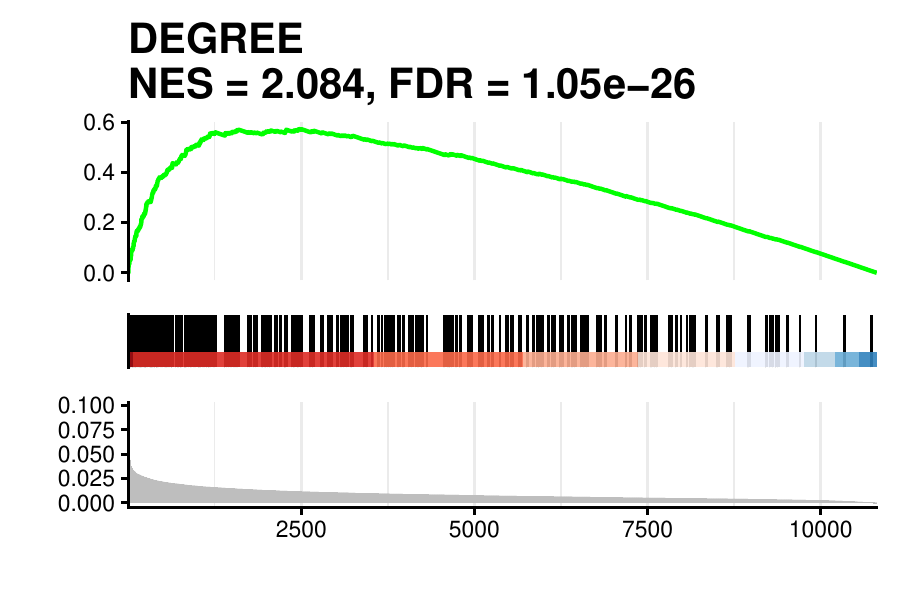}}\\[-5pt]

    {\includegraphics[width=0.40\textwidth]{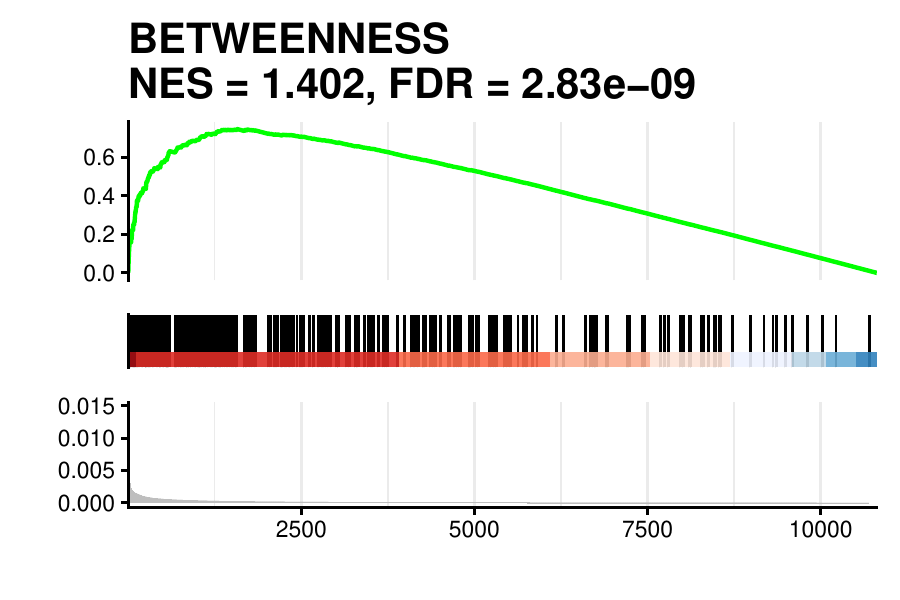}}
    {\includegraphics[width=0.40\textwidth]{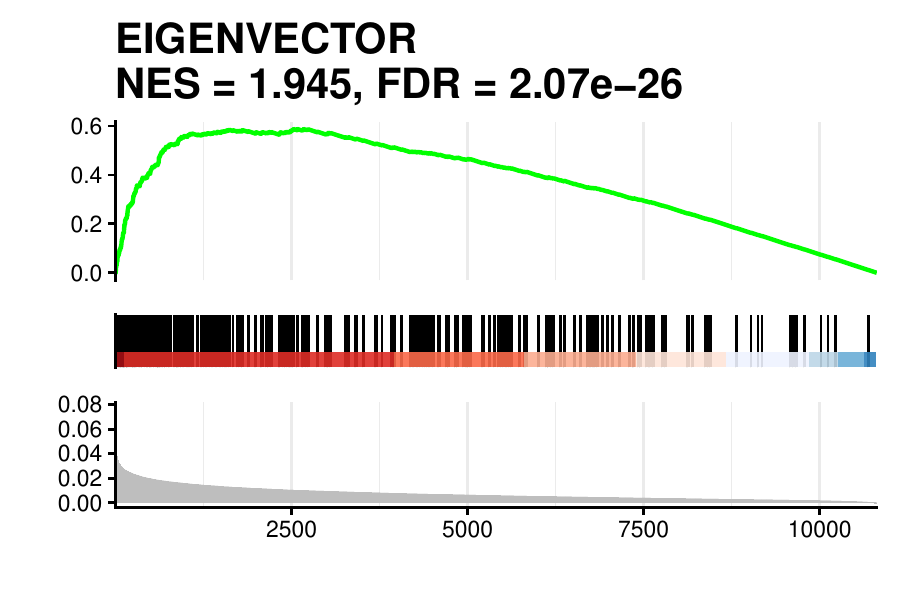}}\\[-5pt]

    {\includegraphics[width=0.40\textwidth]{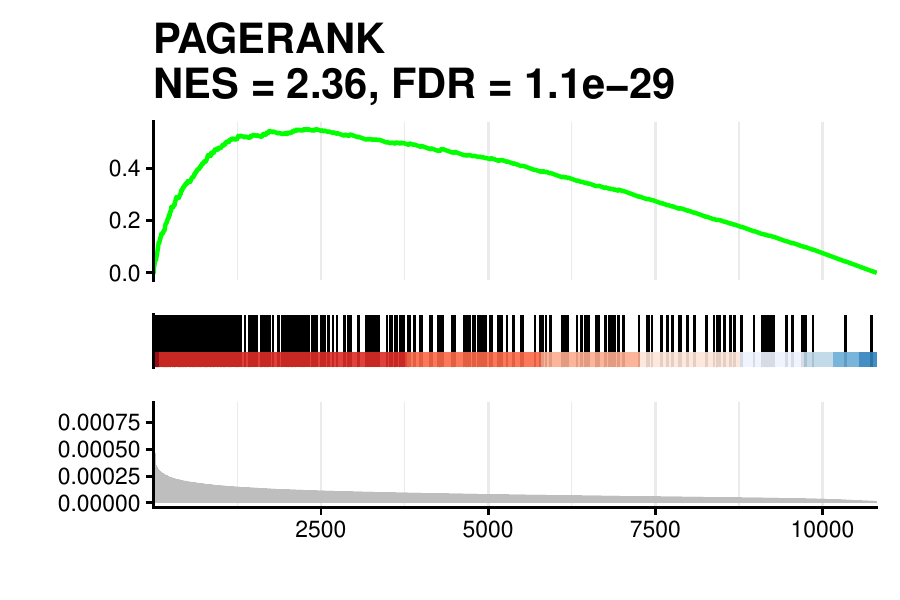}}
    {\includegraphics[width=0.40\textwidth]{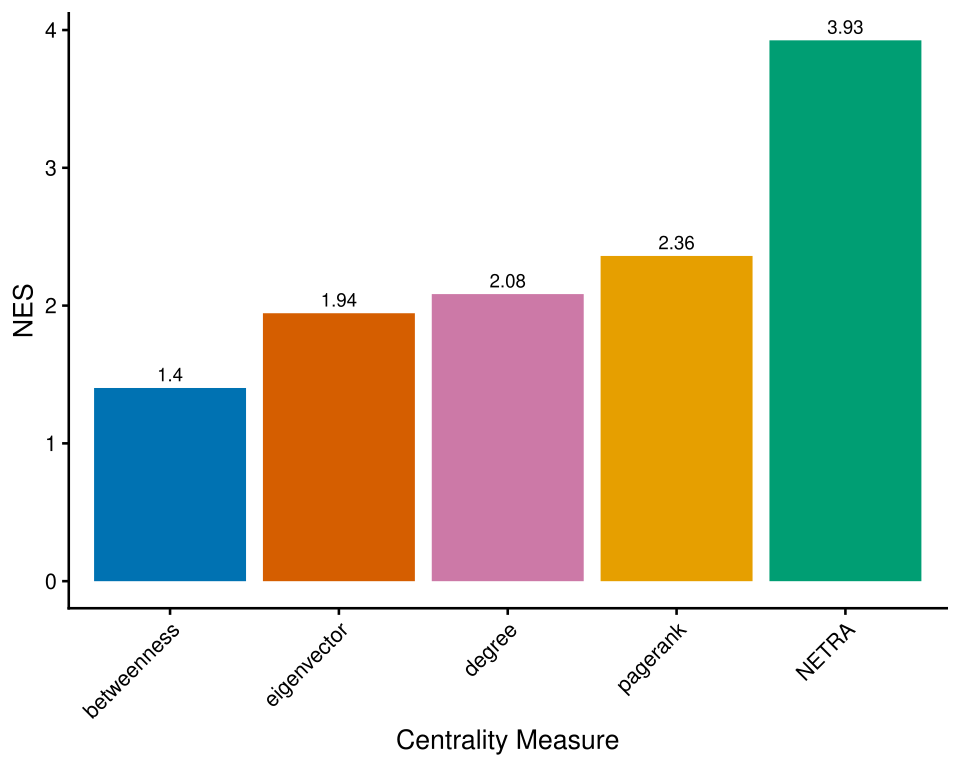}}

    \caption{\textbf{GSEA enrichment profiles and NES comparison for the Alzheimer's disease pathway (KEGG: hsa05010).}
    Panels (a)--(e) show the running enrichment score (green curve), gene rank positions (black bars), and score distribution (gray area) for ranking methods:
    (a) attention-based score NETRA,
    (b) degree,
    (c) betweenness,
    (d) eigenvector, and
    (e) PageRank centrality.
    Panel (f) compares the corresponding normalized enrichment scores (NES) across all five methods.
    The attention-based log2 score NETRA achieves the highest enrichment (NES $\approx$ 3.9, FDR $\approx 1.54\times10^{-28}$), demonstrating superior biological signal integration compared to topology-only centrality measures.}
    \label{fig:gsea_comparison}
\end{figure*}

\inlinehead{Comparative Performance Across Centrality Measures.}
Fig.~\ref{fig:gsea_comparison} presents GSEA enrichment profiles for the Alzheimer's disease pathway (KEGG: hsa05010) across all ranking methods. Attention-based ranking \emph{NETRA} achieved the highest normalized enrichment score (NES $\approx$ 3.9), substantially outperforming all network topology measures: PageRank (NES $\approx$ 2.36), degree (NES $\approx$ 2.08), eigenvector (NES $\approx$ 1.95), and betweenness (NES $\approx$ 1.4) (see Fig. \ref{fig:gsea_comparison} a-e). Only pagerank achieves a slightly more significant FDR than \emph{NETRA}.

\inlinehead{Comparison of GSEA on SIR Result with the Proposed Method.}
GSEA performed on SIR model results yields Ribosome (KEGG: hsa03010) as the top-ranked pathway with a NES of 1.318 ($\text{FDR} = 1.25 \times 10^{-4}$), followed by Coronavirus disease--COVID-19 (KEGG: hsa05171; NES = 1.220, $\text{FDR} = 2.00 \times 10^{-3}$) and Neuroactive ligand-receptor interaction (KEGG: hsa04080; NES = 1.211, $\text{FDR} = 3.83 \times 10^{-3}$). Among the neurodegenerative disease pathways, SIR recovers Parkinson disease (KEGG: hsa05012; NES = 1.184, $\text{FDR} = 1.41 \times 10^{-2}$), Prion disease (KEGG: hsa05020; NES = 1.168, $\text{FDR} = 5.03 \times 10^{-2}$), and Huntington disease (KEGG: hsa05016; NES = 1.152, $\text{FDR} = 5.36 \times 10^{-2}$). Notably, Alzheimer's disease (KEGG: hsa05010) is entirely absent from the SIR-derived enrichment results, whereas it emerges as a prominently enriched pathway under our proposed NETRA score. Furthermore, each of the neurodegenerative pathways recovered by SIR exhibits a lower NES compared to Alzheimer's disease in the NETRA-based analysis, and both Prion disease and Huntington disease fail to reach statistical significance after multiple-testing correction ($\text{FDR} > 0.05$). These findings suggest that the NETRA scoring scheme captures disease-relevant biological signal---particularly for neurodegeneration-associated pathways---that the conventional SIR model fails to prioritize.

\begin{figure*}[!t]
    \centering

    {\includegraphics[width=0.47\textwidth, height=14.2cm]{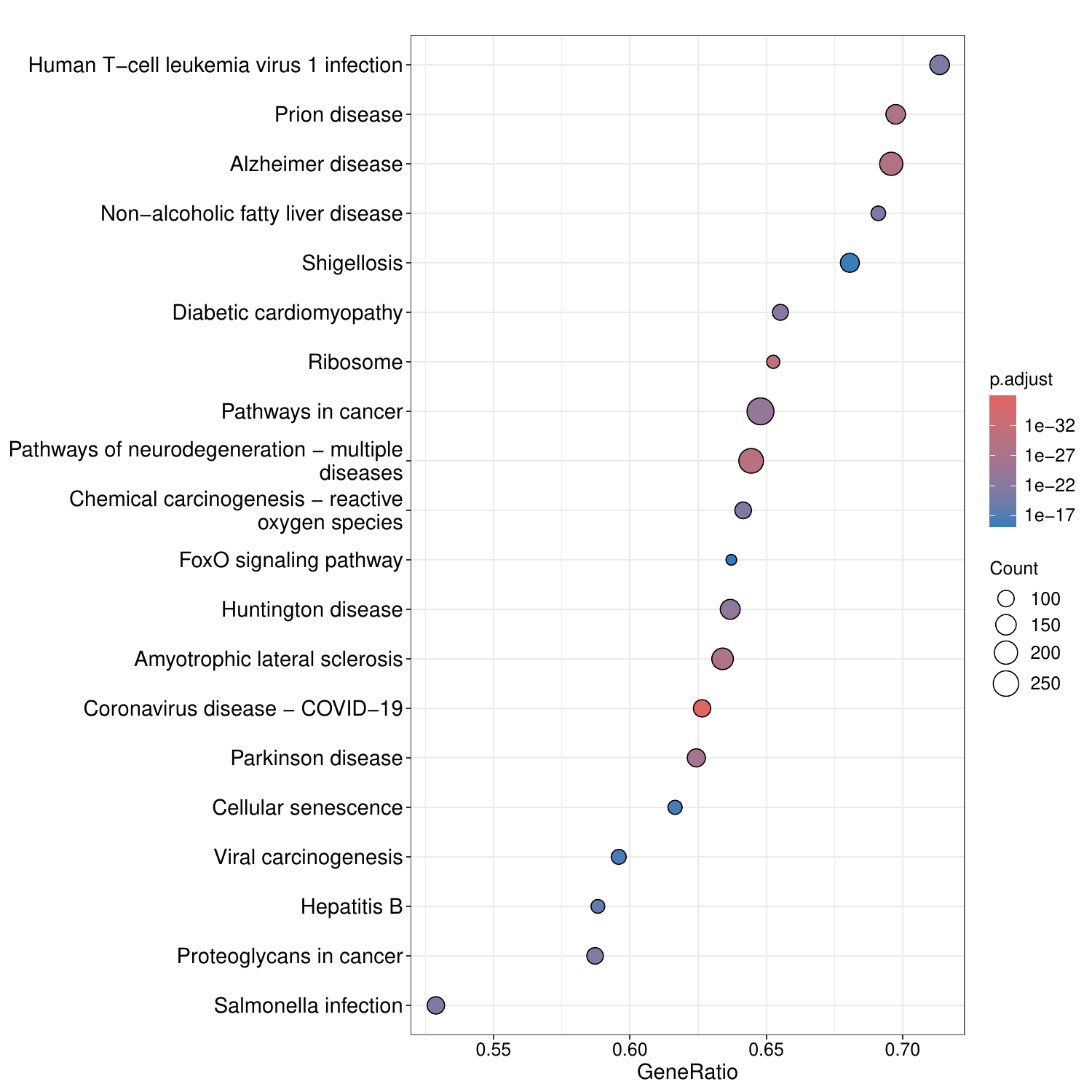}}\hspace{0.7cm}
    {\includegraphics[width=0.47\textwidth, height=14cm]{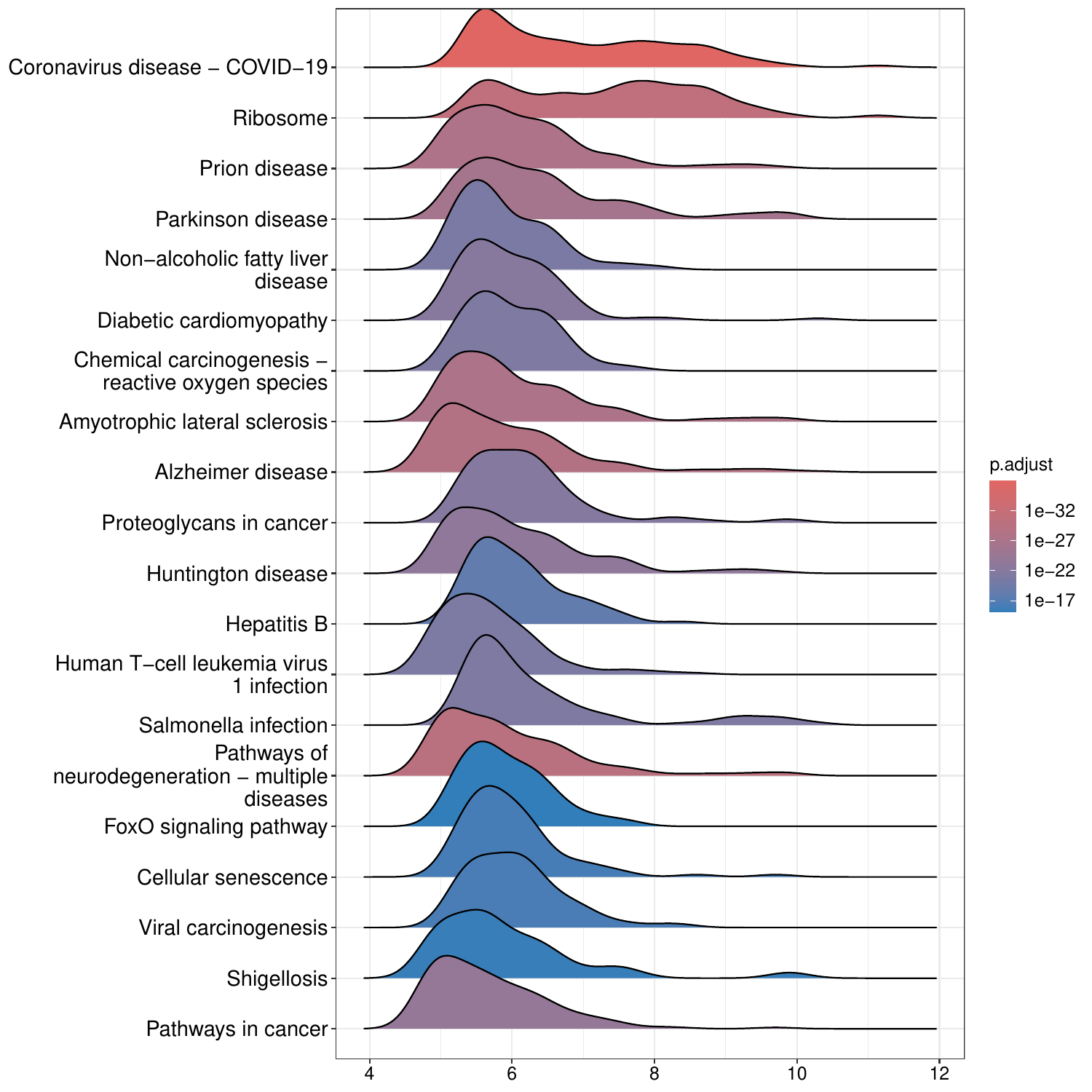}}\\
    \caption{{\bf KEGG pathway enrichment analysis based on attention-derived gene ranking.} \emph{(a) Dot plot summarizing Gene Set Enrichment Analysis (GSEA) results, where dot size represents the gene ratio and color intensity indicates the adjusted p-value. Significant enrichment is observed across multiple neurodegenerative pathways, along with metabolic and synaptic signaling processes, supporting the biological relevance of the proposed attention-based gene prioritization. (b) Ridge plot illustrating the distribution of attention-ranked genes across enriched KEGG pathways. Each ridge represents the density of gene rankings within a pathway, with peak positions indicating the concentration of pathway genes along the ranked list. Consistent early enrichment of neurodegenerative pathways highlights robust prioritization of disease-relevant genes.}}
    \label{fig:gsea_dotplot_ridgeplot}
\end{figure*}

\inlinehead{Broad Pathway Enrichment Landscape.}
Beyond Alzheimer's disease, attention-based ranking enriched multiple neurodegenerative pathways with high significance. Figure~\ref{fig:gsea_dotplot_ridgeplot} presents the top 20 enriched KEGG pathways, dominated by neurodegeneration-related processes: Pathways of neurodegeneration - multiple diseases (FDR = 1.43×10$^{-30}$), Parkinson's disease (FDR = 4.94×10$^{-27}$), Huntington's disease (FDR = 6.07×10$^{-24}$), Amyotrophic lateral sclerosis (FDR = 2.85×10$^{-28}$), and Prion disease (FDR = 1.65×10$^{-28}$). The enrichment of protein synthesis machinery (Ribosome, FDR = 1.55×10$^{-31}$) and proteostasis-related pathways indicates the model captures cellular stress mechanisms common to neurodegenerative diseases. The ridge plot (Figure~\ref{fig:gsea_dotplot_ridgeplot}) visualizes the distribution of gene rankings within each pathway, showing consistent enrichment patterns across neurodegenerative processes.

Notably, cancer-related pathways (Pathways in cancer, Proteoglycans in cancer, Viral carcinogenesis) and infectious disease responses (Coronavirus disease - COVID-19, Hepatitis B, Salmonella infection) also achieved significant enrichment. This reflects shared molecular mechanisms between neurodegeneration and cancer, particularly, dysregulated cell cycle control, apoptosis, and proteostasis, as well as immune and inflammatory responses central to both infectious and neurodegenerative pathologies.

\subsubsection{Enrichment Analyses of Top-Ranked Genes}
\begin{figure*}[!t]
    \centering

    \includegraphics[width=0.32\textwidth, height=7.40cm]{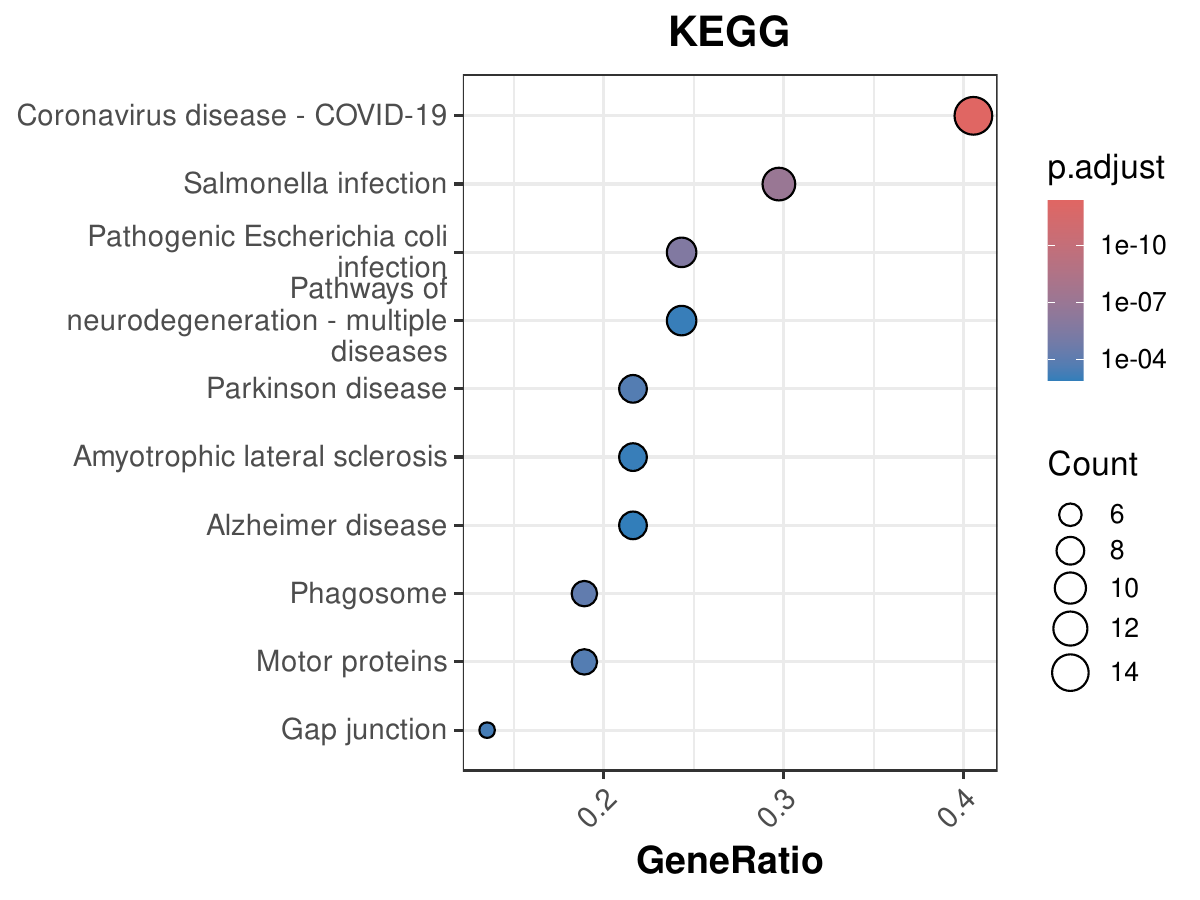}\label{fig:kegg_pea}
    {\includegraphics[width=0.32\textwidth, height=7.40cm]{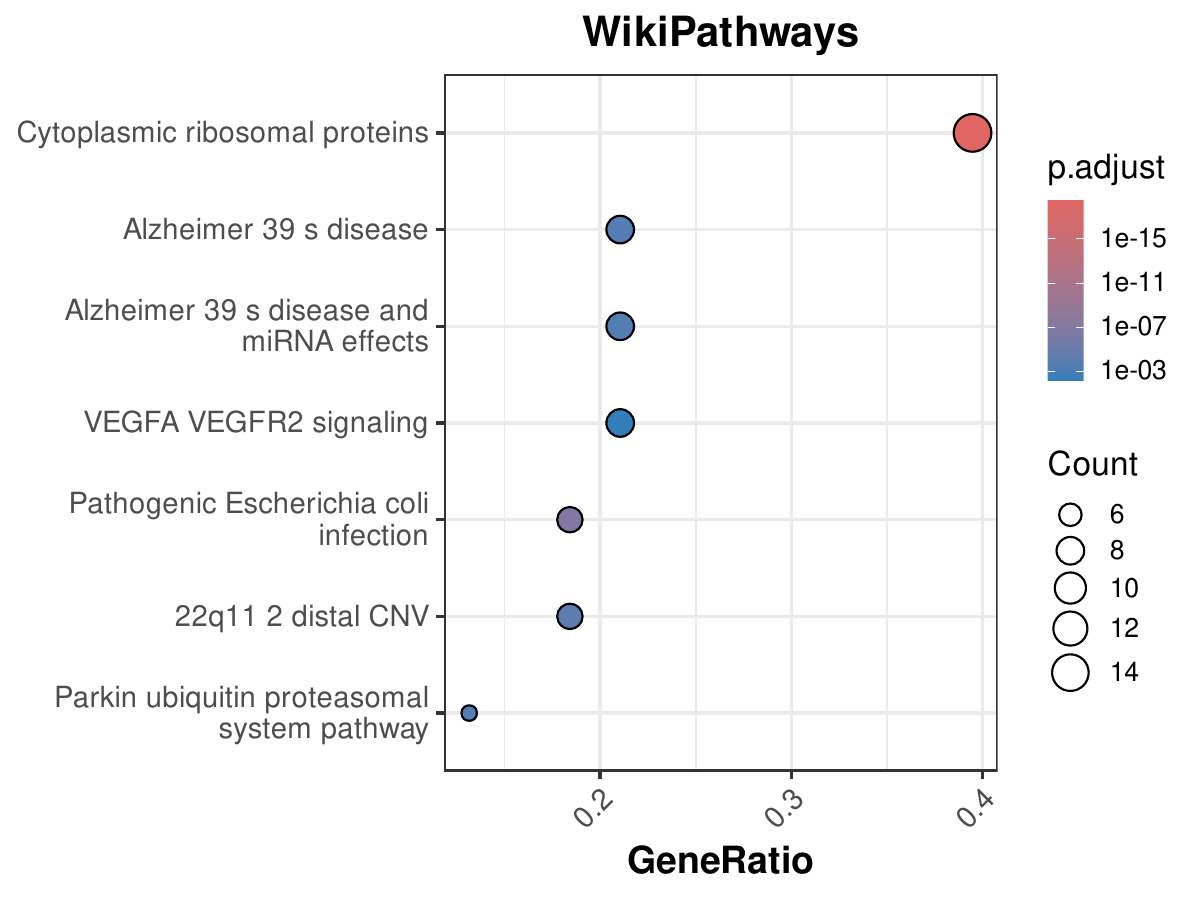}}\label{fig:wp_pea}    
    {\includegraphics[width=0.32\textwidth, height=7.40cm]{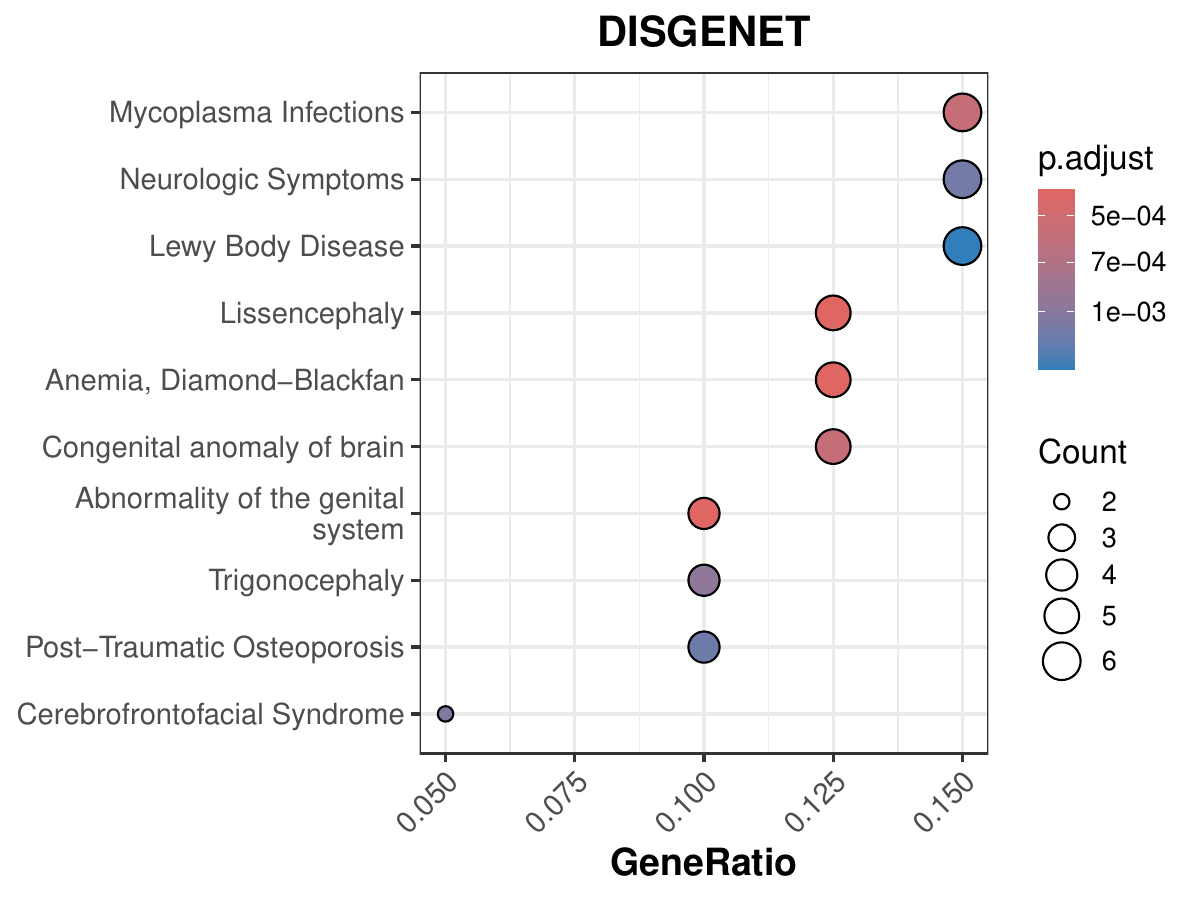}}\label{fig:disgenet_enrichment}
    \vspace{-0.95cm}

    \label{fig:chr12_q13}{\includegraphics[width=0.9\textwidth]{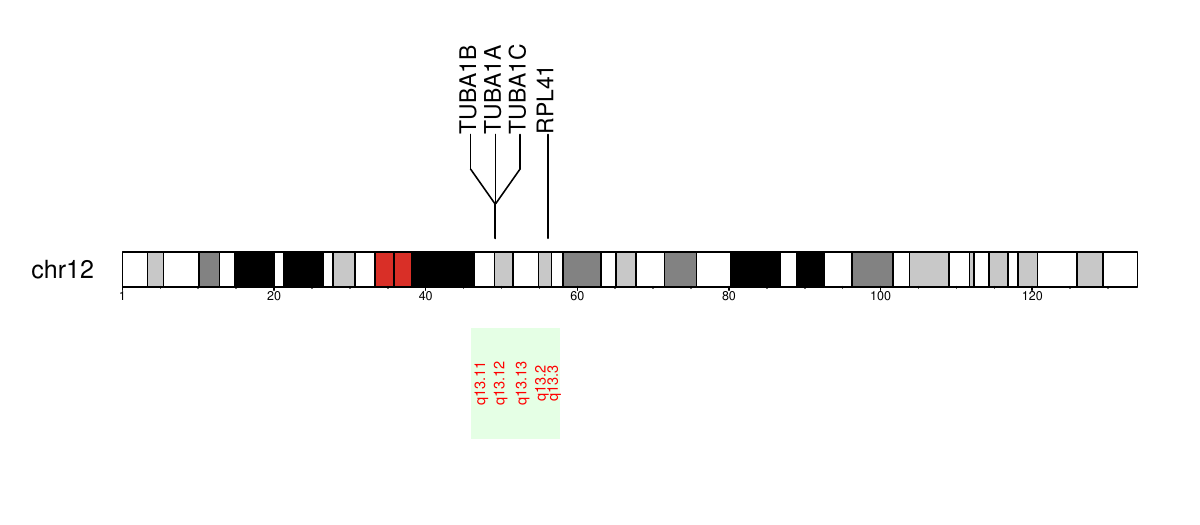}}
    \caption{{\bf Multi-level functional and genomic validation of attention-prioritized genes.} \emph{(a)KEGG pathway enrichment analysis of the top 40 attention-ranked genes, showing significant overrepresentation of neurodegeneration-related pathways alongside apoptotic and iron metabolism processes.
 (b)WikiPathways enrichment analysis of the top 40 genes, highlighting Alzheimer’s disease–associated signaling pathways and shared molecular mechanisms relevant to neurodegeneration.
 (c)DISGENET disease enrichment analysis for the top 40 genes, revealing convergence across multiple neurodegenerative and neurological disorders, indicating shared disease mechanisms.
 (d) Genomic localization of attention-prioritized genes on chromosome 12, highlighting a cluster of four genes mapping to the 12q13 cytoband, a locus previously implicated in Alzheimer’s disease through genome-wide association studies \cite{beecham_12q13_2009, yu_12q13_2011}}}\label{fig:multi_validate}

\end{figure*}

We performed targeted enrichment of the top 40 attention-ranked genes to validate specific biological mechanisms (Fig. \ref{fig:multi_validate}).

\inlinehead{Chromosome Region Enrichment.}
Chromosome region enrichment identified chr12q13, supported by four genes (Fig.~\ref{fig:multi_validate}-d)) . This locus has been implicated in late-onset Alzheimer's disease \cite{beecham_12q13_2009, yu_12q13_2011}, validating that attention weights capture genetically relevant disease architecture.

\inlinehead{Pathway Enrichment.}
KEGG and WikiPathways enrichment (Fig.~\ref{fig:multi_validate}- a,b) revealed convergent mechanisms. KEGG pathways emphasized neurodegeneration (Pathways of neurodegeneration - multiple diseases, Alzheimer disease, Parkinson disease, Amyotrophic lateral sclerosis), infection response (Coronavirus disease - COVID-19, Salmonella infection, Pathogenic Escherichia coli infection), and cellular processes (Phagosome, Motor proteins, Gap junction). WikiPathways highlighted AD-specific signatures (Alzheimer's disease, Alzheimer's disease and miRNA effects), protein degradation (Parkin ubiquitin proteasomal system pathway), angiogenesis (VEGFA-VEGFR2 signaling), and protein synthesis (Cytoplasmic ribosomal proteins). Both databases converge on neurodegeneration and infection response pathways, validating attention-based prioritization across complementary annotation systems.

\inlinehead{Disease Enrichment.}
DISGENET enrichment (Fig.~\ref{fig:multi_validate}-c) revealed neurological conditions including Lewy Body Disease, Neurologic Symptoms, and developmental disorders (Lissencephaly, Trigonocephaly, Congenital anomaly of brain). Additional enriched terms include Anemia Diamond-Blackfan, Cerebrofrontofacial Syndrome, and Post-Traumatic Osteoporosis, reflecting diverse biological processes captured by attention-prioritized genes.

\subsubsection{Cross-Pathway Gene Conservation in Neurodegeneration}
To investigate why multiple neurodegenerative pathways show concurrent enrichment, we examined the overlap of core enrichment genes across Alzheimer's disease, Parkinson's disease, Huntington's disease, Amyotrophic lateral sclerosis, and Prion disease pathways.

Our attention-ranked gene list shows co-enrichment of these neurodegenerative pathways driven by shared core genes, particularly tubulin family members (TUBA1B, TUBA1A, TUBA1C, TUBB2A, TUBB) central to axonal transport and cytoskeletal integrity.

\begin{figure*}[!htbp]
    \centering
    \includegraphics[width=0.90\textwidth]{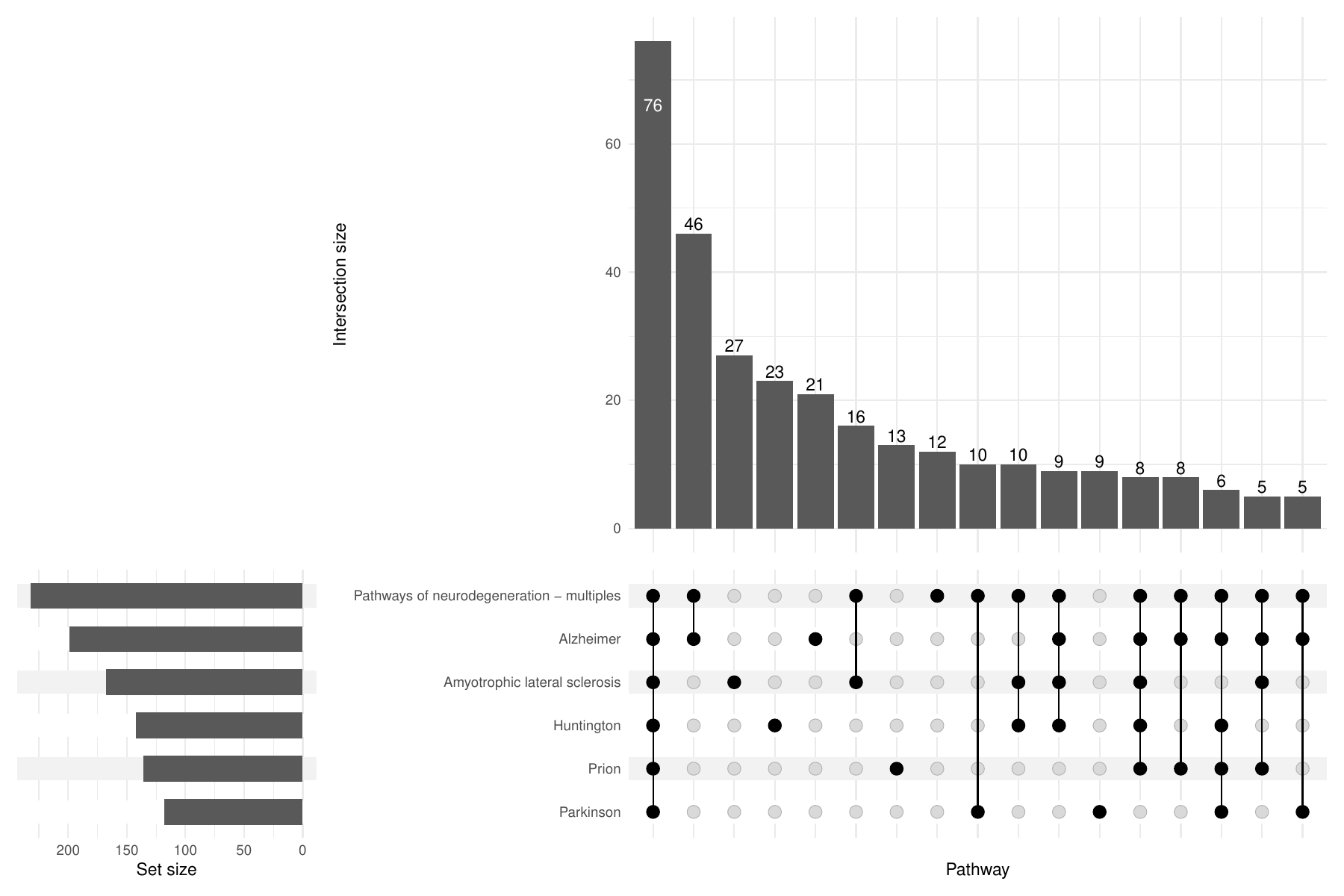}
    \caption{\textbf{UpSet plot showing shared genes across neurodegenerative pathways.} Bar heights indicate the number of genes in each intersection. The largest intersections reveal core neurodegeneration genes prioritized by attention scores.}
    \label{fig:neuro_upset}
\end{figure*}

\begin{figure*}[!htbp]
    \centering
    \includegraphics[width=0.75\textwidth]{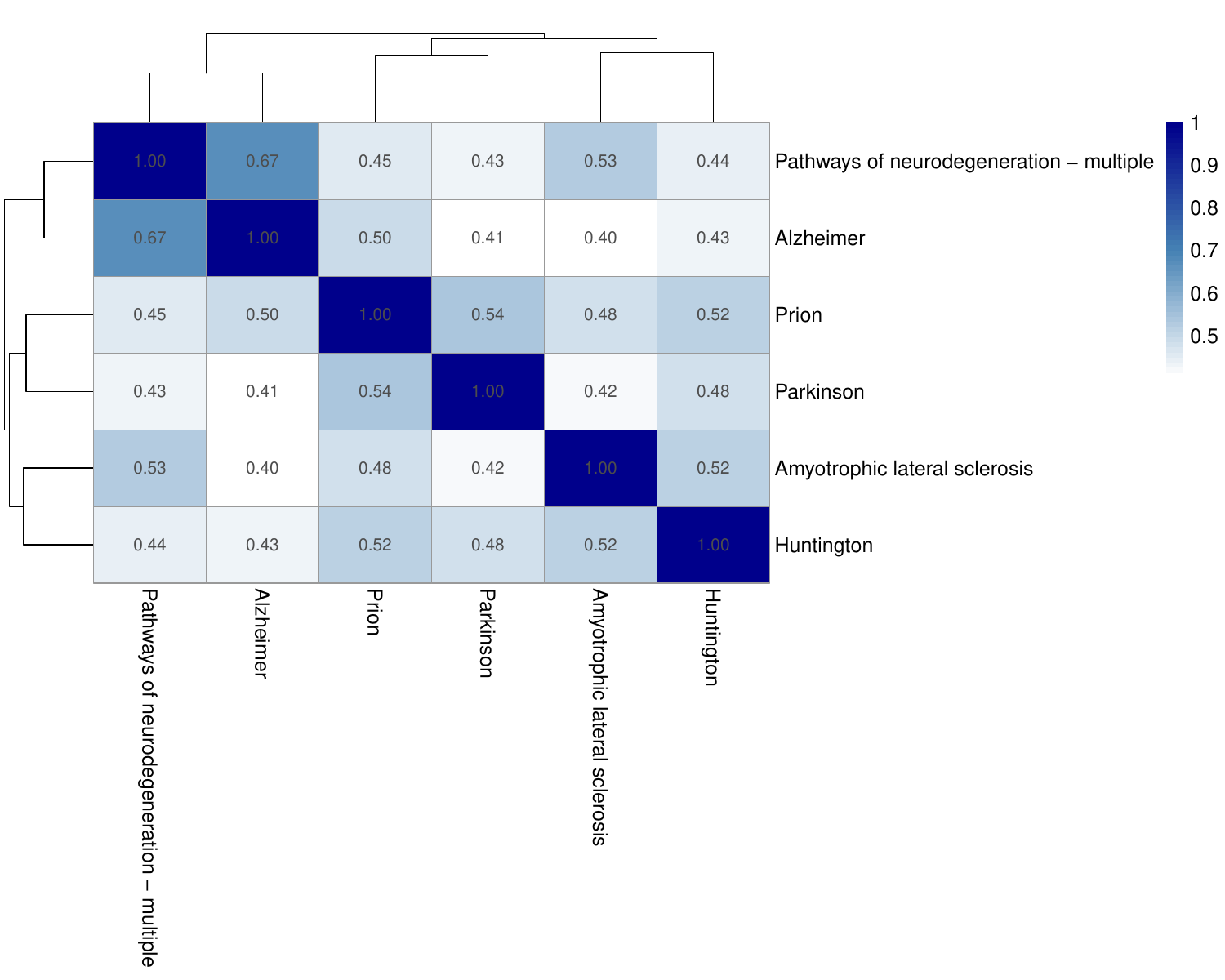}
    \caption{\textbf{Jaccard similarity between neurodegenerative pathway gene sets.} Higher values indicate greater overlap in core enrichment genes.}
    \label{fig:neuro_jaccard}
\end{figure*}

Figure~\ref{fig:neuro_upset} presents the intersection patterns across pathways, revealing [X] genes shared across all five conditions. The Jaccard similarity heatmap (Figure~\ref{fig:neuro_jaccard}) quantifies pairwise pathway overlap, with highest similarity observed between Alzheimer's disease and Prion disease (Jaccard = 0.54).

These findings suggest that attention-based prioritization captures conserved neurodegeneration machinery rather than pathway-specific noise, validating the biological coherence of the learned gene rankings.

\section{Conclusion}
In this study, we presented an attention-based multimodal framework for prioritizing Alzheimer’s disease–associated genes by integrating heterogeneous biological evidence across multiple data modalities. By combining gene regulatory networks derived from various transcriptomic data and auxiliary biological networks. By combining gene representations learned through VAEs with inferred graph-based embeddings extracted via a BERT model and Graph Transformer, the proposed approach named as \emph{NETRA} moves beyond traditional network-centric gene prioritization strategies. 

Our results demonstrate that the learned gene embeddings exhibit clear structural organization, with well-defined communities in the embedding space. Attention-prioritized genes are distributed across multiple embedding clusters and form densely connected interaction modules, indicating that the model captures diverse but functionally coherent biological processes rather than trivially selecting network hubs. Importantly, the generated network preserves key topological properties of the input ensemble, including clustering behavior, global efficiency, and heavy-tailed degree distributions, supporting the structural validity of the learned graph.

Functional validation further highlights the strength of attention-based prioritization. Compared to classical centrality measures, attention-derived rankings consistently achieved stronger enrichment of Alzheimer’s disease pathways and related neurodegenerative processes. Enrichment analyses revealed convergence on conserved neurodegeneration mechanisms, including cytoskeletal organization, proteostasis, and axonal transport, while also capturing shared molecular signatures across multiple neurological disorders. Targeted enrichment of top-ranked genes further identified disease-relevant pathways, chromosomal loci, and known AD-associated regions, reinforcing the biological relevance of the attention scores.

Overall, these findings provide compelling evidence that graph attention mechanisms offer a principled and multimodal-aware strategy for disease gene prioritization. By jointly modeling network structure, gene expression dynamics, and auxiliary biological knowledge, the proposed framework improves interpretability and biological coherence relative to traditional network-based methods. While the present study focuses on Alzheimer’s disease as a case study, the framework itself is not disease-specific and is readily generalizable to other complex disorders. In this sense, the proposed approach can serve as a flexible alternative to classical centrality-based prioritization strategies, offering a more expressive and context-aware paradigm for disease gene discovery across diverse biological settings.
\FloatBarrier
\clearpage

\section*{Funding}
The research work is supported by the Department of Biotechnology (DBT), GoI, under the project BT/PR51150/NER/95/1996/2023. The work is also partially supported by IDEAS-TIH, ISI-Kolkata.


\bibliographystyle{unsrt}{}
\bibliography{bibliography/reference}{}

\begin{IEEEbiography}[{\includegraphics[width=1in,height=1.25in]{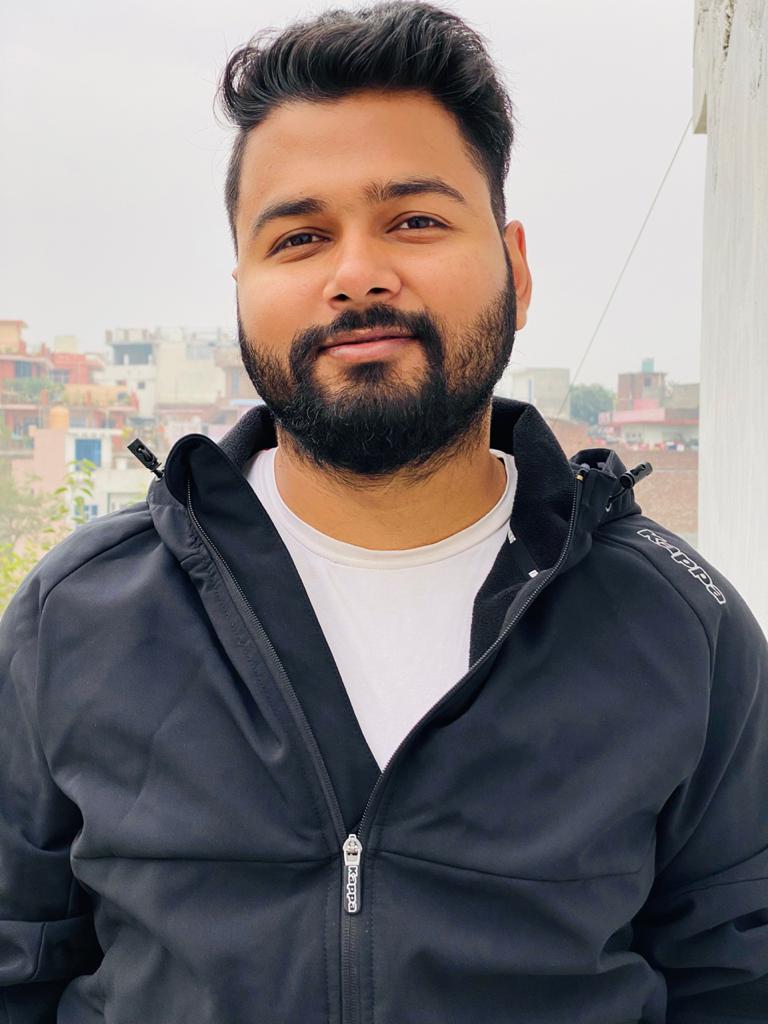}}]{Binon Teji} is a Senior Research Fellow at Network Reconstruction and Analysis Lab and Ph.D.  scholar at Sikkim (Central) University, Gangtok, India. He received his Master of Computer Application degree from the University of Jammu, Jammu and Kashmir, India. His research areas includes Graph Representation Learning, Computational Biology.
\end{IEEEbiography}
\vspace{-0.5in}

\begin{IEEEbiography}[{\includegraphics[width=1in,height=1.25in]{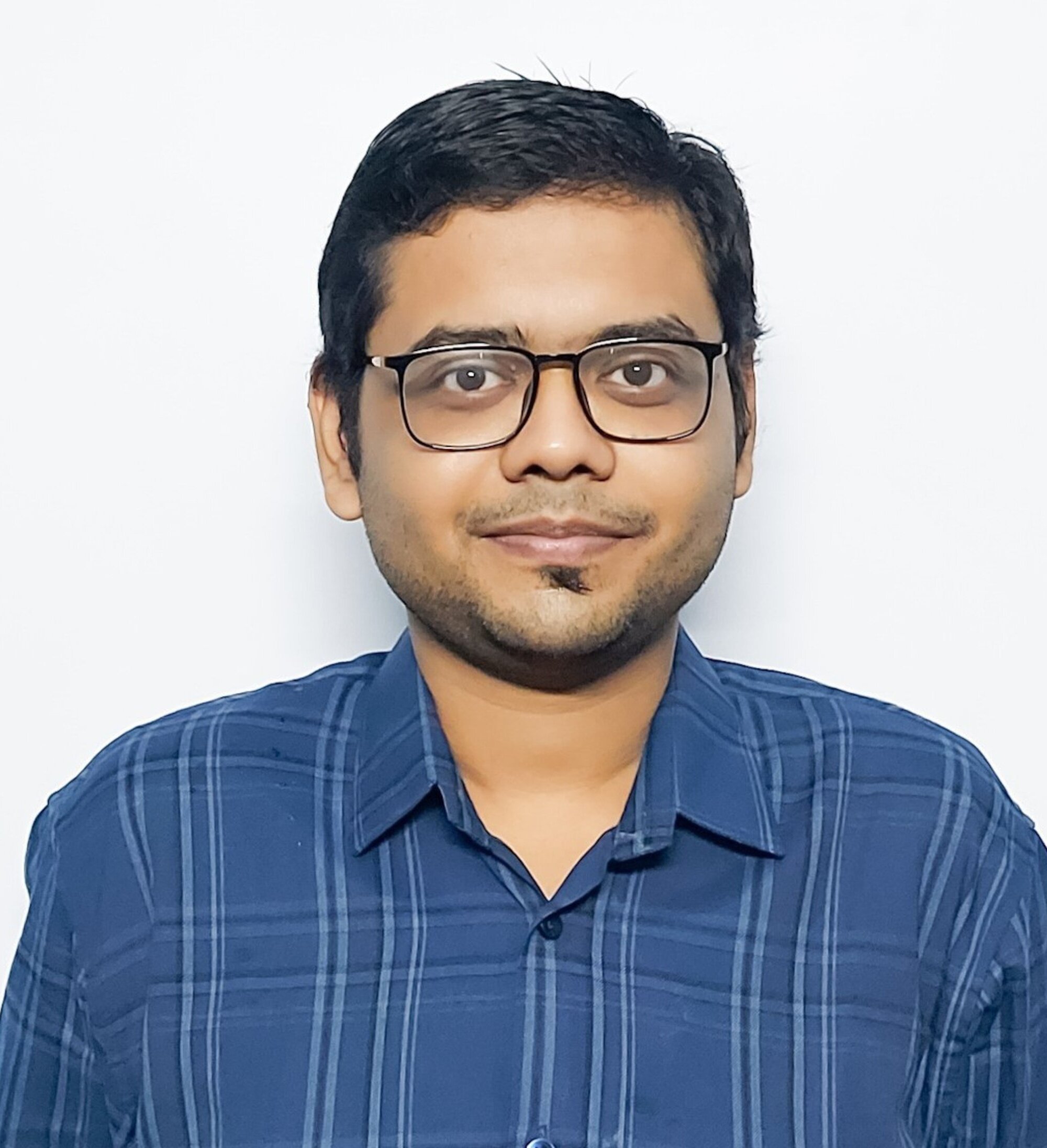}}]{Subhajit Bandyopadhyay} is a Project Fellow at Network Reconstruction and Analysis Lab at Sikkim (Central) University, Gangtok, India. He received his Ph.D. (Mathematics) degree from Tezpur University, Assam, India. His research areas includes high-dimensional data, topological data analysis, computational biology, and combinatorics.
\end{IEEEbiography}
\vspace{-0.5in}

\begin{IEEEbiography}[{\includegraphics[width=1in,height=1.25in]{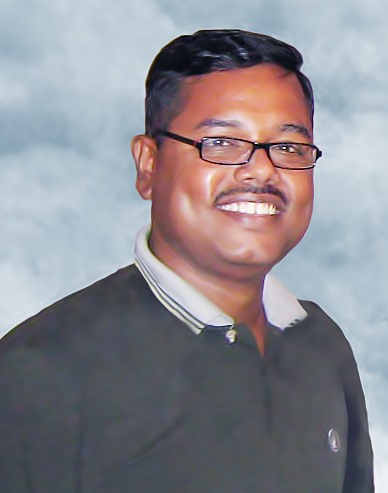}}]{Swarup Roy} is a Professor in the Department of Computer Science \& Engineering at Tezpur University. His research interests include graph machine learning in computational biology and health informatics. He has published in reputed international journals and conferences, written books, and serves as associate editor and on technical committees of several international journals and conferences.
\end{IEEEbiography}

\end{document}